\pdfoutput=1
\documentclass{article}
\usepackage[eandd,final]{neurips_2026}

\usepackage{amsmath,amsfonts,bm}

\def\eqref#1{equation~\ref{#1}}

\def\1{\bm{1}}

\DeclareMathAlphabet{\mathsfit}{\encodingdefault}{\sfdefault}{m}{sl}
\SetMathAlphabet{\mathsfit}{bold}{\encodingdefault}{\sfdefault}{bx}{n}

\usepackage[utf8]{inputenc}
\usepackage[T1]{fontenc}
\usepackage{hyperref}
\usepackage{url}
\usepackage{booktabs}
\usepackage{needspace}
\usepackage{longtable}
\usepackage{amsfonts}
\usepackage{nicefrac}
\usepackage{microtype}
\usepackage{graphicx}
\usepackage{array}
\usepackage{tabularx}
\usepackage{amsmath}
\usepackage{amssymb}
\usepackage{subcaption}
\usepackage[dvipsnames]{xcolor}
\usepackage{placeins}
\usepackage{float}
\usepackage[skins]{tcolorbox}

\definecolor{analogyaccent}{RGB}{45, 110, 120}
\newtcolorbox{analogybox}[1][]{%
  enhanced,
  colback=white,
  colframe=analogyaccent,
  colbacktitle=analogyaccent,
  coltitle=white,
  fonttitle=\bfseries\small,
  title={#1},
  attach boxed title to top left={xshift=5mm, yshift=-2.5mm},
  boxed title style={colframe=analogyaccent, boxrule=0pt, arc=1.5mm},
  arc=1.5mm,
  boxrule=0.5pt,
  left=3mm, right=3mm, top=3mm, bottom=2.5mm,
  toptitle=0.8mm, bottomtitle=0.8mm,
  before skip=6pt,
  after skip=6pt,
}

\newcommand{\papertitle}{Reset Dependence Is Embodied: Training Protocol Changes Which Morphologies Appear Optimal}

\hypersetup{
  colorlinks=true,
  allcolors=blue!60!black,
  pdfborder={0 0 0},
  breaklinks=true,
  pdftitle={\papertitle},
  pdfauthor={Kevin Qiu, Zhao Yang, Marek Cygan},
}

\newcommand{\codeurl}{\url{https://github.com/kevinxqiu/reset-dependence}}
\newcommand{\pageurl}{\href{https://reset-dependence.github.io}{\nolinkurl{reset-dependence.github.io}}}

\newcommand{\appcell}[1]{\raggedright\arraybackslash #1}

\newcolumntype{L}[1]{>{\raggedright\arraybackslash}p{#1}}
\newcolumntype{Y}{>{\raggedright\arraybackslash}X}

\title{Reset Dependence Is Embodied: Training Protocol\\Changes Which Morphologies Appear Optimal}

\author{%
  Kevin Qiu$^{1}$\quad Zhao Yang$^{2}$\quad Marek Cygan$^{1}$\\[3pt]
  {\normalfont $^{1}$University of Warsaw\quad $^{2}$Vrije Universiteit Amsterdam}\\[2pt]
  {\normalfont \texttt{kevinxqiu@gmail.com}}
}

\newcommand{\apptochtarget}[2]{%
  \hyperref[#2]{#1}}

\newcommand{\apptocmain}[2]{%
  \vspace{3.5pt}%
  \noindent{\bfseries\apptochtarget{#1}{#2}}\dotfill{\bfseries\apptochtarget{\pageref*{#2}}{#2}}\par\vspace{-1pt}}
\newcommand{\apptocsub}[2]{%
  \noindent\hspace{1.6em}\apptochtarget{#1}{#2}\dotfill\apptochtarget{\pageref*{#2}}{#2}\par\vspace{-2pt}}

\begin{document}

\maketitle

\begin{abstract}
Robot co-design benchmarks rank bodies after training their controllers under a fixed episode protocol, but that protocol can change the ranking itself. Consequently, two studies can select different bodies from the same corpus even when they use the same task and evaluator. We study one part of that protocol, the reset rule, which ends a training episode when the robot's torso falls below a height threshold. We train PPO controllers for 100 DERL-evolved UNIMAL bodies with this threshold at 20\% of each body's starting torso height (Standard), at 50\% (Strict), or with no threshold (Reset-free), and score every controller with the same evaluator. The threshold changes which bodies look best: the rankings after Strict and Reset-free training correlate at only 0.21, against about 0.89 if seed noise were the only difference. Removing fall resets costs some bodies far more than others. Bodies evolved on flat ground lose 45\% of their per-step reward relative to Standard training, and those evolved on variable terrain only 16\%. Permutation tests over bodies support this difference, and on 24 bodies, flat-ground bodies gain the most from Strict over Reset-free training under both PPO and SAC. Choosing bodies by their worst reward across the three protocols costs little: in sample, they keep at least 91\% of the reward of each protocol's best bodies, and over 80\% averaged across held-out seeds. Co-design benchmarks should state their reset rule, test whether their rankings depend on it, and report bodies that stay strong across protocols. Project page: \pageurl{}.
\end{abstract}

\begin{figure}[H]
    \centering
    \vspace{-0.75em}
    \includegraphics[width=\textwidth]{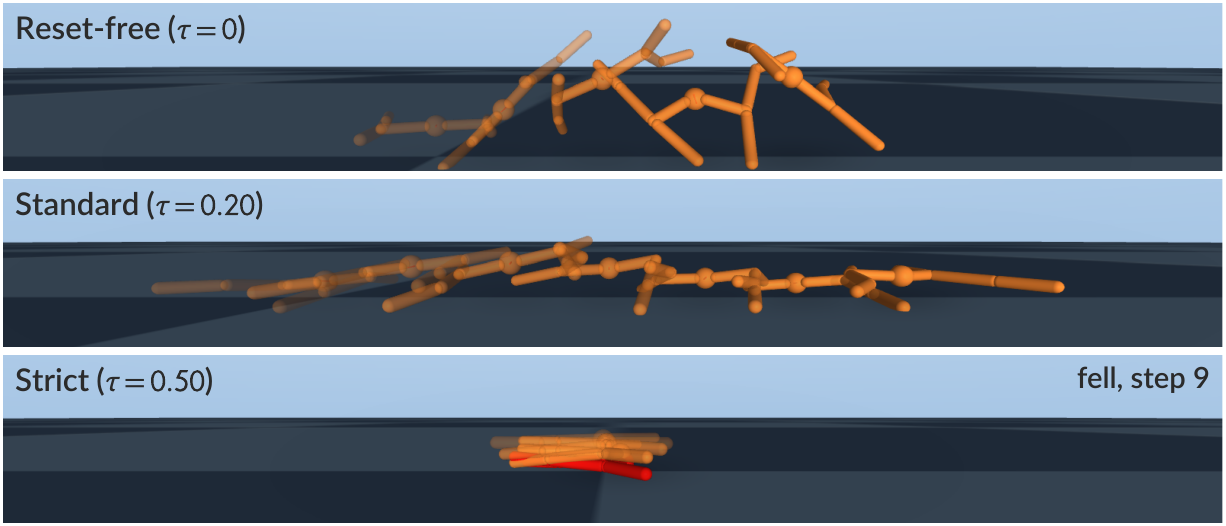}
    \vspace{-0.35em}
    \caption{\textbf{Reset dependence is embodied.} In illustrative retrains, \emph{Floor A} walks under Standard ($\tau{=}0.20$), makes no forward progress under Reset-free ($\tau{=}0$), and its Strict ($\tau{=}0.50$) episode ends at step~9. \textbf{Takeaway:} controllers of one body can behave very differently.}
    \label{fig:gait_chronophoto_hero}
    \vspace{-0.4em}
\end{figure}

\section{Introduction}
\label{sec:intro}

Robot co-design benchmarks rank bodies after training their controllers under a fixed episode protocol, but that protocol can change the ranking itself. Consequently, two studies can select different bodies from the same corpus even when they use the same task and evaluator. Co-design methods search over bodies, and recent studies also vary the task, but each commits to a single recipe for training the controllers that score the bodies~\citep{gupta2021embodied, lu2025bodygen, ai2025embodiment}. Such training and evaluation protocols are known to shape the learned controller~\citep{henderson2018deep, pardo2018time, machado2018revisiting}. We ask whether they also change which \emph{body} appears optimal.

We study one part of the protocol, the \emph{reset rule}. During training, an episode ends when the robot's torso falls below a height threshold, and the robot starts again from its initial pose. Real robots often cannot rely on cheap fall resets~\citep{sharma2022autonomous, eysenbach2018leave}, so a body selected under reset-rich training may be optimized for resets that are unavailable at deployment.

\begin{analogybox}[Marathon Runner vs.\ Sprinter]
A sprinter and a marathon runner are both elite athletes, but they are not interchangeable. In a 100 m race, a single stumble often decides the outcome, whereas in a marathon, the ability to recover matters. The event, not anatomy alone, determines which body wins. Robot morphology co-design follows the same logic. A reset rule determines whether a fall ends an episode or becomes a state the controller must learn to handle. Changing this rule shifts which morphology appears optimal. Our experiments measure the resulting ranking shift.
\end{analogybox}

We test this on 100 bodies from the UNIMAL design space, evolved by Deep Evolutionary Reinforcement Learning (DERL)~\citep{gupta2021embodied} in three families: on flat ground (Floor), on variable terrain (VT), and for pushing a box across variable terrain (MVT). Here every controller is trained from scratch on the same task, walking forward on flat ground. We compare three training protocols that set the reset threshold to a fraction $\tau$ of the body's starting torso height: Standard ($\tau{=}0.20$), Strict ($\tau{=}0.50$), and Reset-free ($\tau{=}0$), which never resets after a fall. Each protocol is run with three seeds, and every other episode setting is specified identically. We change the rule only during training and score every controller with the same evaluator. Our contributions are:

\begin{enumerate}
    \item \textbf{The reset rule is part of the benchmark specification.} Changing the reset rule during training reorders the bodies. After Strict and Reset-free training, the rankings of all $100$ bodies correlate at Spearman $\rho = 0.21$, against about $0.89$ if seed noise were the only difference, and the reordering persists when runs are scored without a cutoff. As a control, retraining 60 bodies with a threshold below the torso's resting height leaves their Reset-free ranking almost unchanged. A morphology ranking is therefore underspecified without the reset rule that produced it.

    \item \textbf{Reset dependence is embodied.} How much a body loses without fall resets differs across evolutionary families. Under proximal policy optimization (PPO), Floor bodies lose $45\%$ of their per-step reward under Reset-free training relative to Standard training, but VT bodies lose only $16\%$. The difference holds in permutation tests over bodies and when runs are scored without a cutoff, and on 24 bodies, Floor bodies gain the most from Strict over Reset-free training under both PPO and soft actor-critic (SAC; Table~\ref{tab:family}, Figure~\ref{fig:hero}, Sections~\ref{sec:cross_algo} and~\ref{sec:distance_audit}).

    \item \textbf{Selecting for robustness costs little reward.} In sample, bodies chosen by their worst per-step reward across the three protocols keep at least 91\% of the reward of each protocol's own best bodies, and over 80\% averaged across held-out seeds (Section~\ref{sec:regret}, Table~\ref{tab:robust_selection}). Section~\ref{sec:audit} turns this into a short audit for benchmark authors that also checks whether the reset rule acts on every body.
\end{enumerate}

Every controller is trained from scratch, so the family difference does not come from controllers inherited from evolution. One explanation is what the bodies were selected for. Floor bodies were evolved for speed on predictable surfaces, so once a fall no longer ends the episode, they must act from states that were rare during their evolution. VT bodies were evolved on rough terrain that exposes them to such states. Our experiments are consistent with this account. The audit applies to any co-design benchmark that scores bodies under a reset rule. We release the code, the records of all 900 runs, and a script that recomputes 2,401 values from the released records on a CPU (Section~\ref{sec:artifacts}).

\section{Related Work}
\label{sec:related}

\paragraph{Morphology co-design.}
Methods that co-optimize morphology and control include evolutionary search~\citep{sims1994evolving, gupta2021embodied, bhatia2021evolution}, differentiable design~\citep{chen2021hardware}, learned design policies~\citep{yuan2022transform2act}, efficient embodiment generators such as BodyGen~\citep{lu2025bodygen}, and LLM-guided morphology design such as RoboMorph~\citep{qiu2024robomorph}. Recent LLM-based co-design also treats the reward itself as a design variable~\citep{qiu2025debate2create}. DERL~\citep{gupta2021embodied} introduced the UNIMAL space used here, MetaMorph~\citep{gupta2022metamorph} showed that universal controllers generalize across morphologies, recent work identifies embodiment scaling laws across about 1,000 bodies~\citep{ai2025embodiment}, and \citet{yue2025robocraft} target humanoid morphology-control co-design for fall recovery. \citet{strgar2026accelerated} accelerate co-design through morphological pretraining.

These methods vary bodies, controllers, tasks, rewards, or design generators, but they generally fix one termination and evaluation protocol instead of auditing termination as a benchmark variable. We hold the reward fixed and show that the reset rule can change which body ranks best. \citet{mertan2025cooptimization} offer a complementary critique: simultaneous brain-body co-optimization can undervalue offspring with newly mutated bodies, so the search can miss strong bodies. We show instead that the training protocol used to score bodies can \emph{redefine} which bodies are best.

\paragraph{Termination in RL.}
Prior work treats termination as part of the task. \citet{pardo2018time} showed that time limits introduce a non-Markov bias and proposed partial-episode bootstrapping. That correction changes how values are bootstrapped at episode boundaries, whereas our intervention changes which fall states the controller sees during training. CaT~\citep{chanesane2024cat} recasts safety constraints as stochastic terminations for legged locomotion, on a fixed robot platform. Constraint- or bootstrapping-based alternatives may change how sensitive a controller is to termination, but they have not been used to audit morphology rankings under different reset rules. \citet{li2023survival} report that offline RL agents develop a survival bias from the data distribution. These works study termination on a fixed task, robot or dataset, whereas in our setting the body matters, because the same threshold leads to qualitatively different outcomes for different bodies.

\paragraph{Reset-free RL and morphology.}
\citet{sharma2022autonomous} formalized autonomous RL, and later work developed reset-free algorithms for fixed embodiments~\citep{patil2024risc, yang2025morefree, gupta2021mtrf}. \citet{powers2020morphology} report that morphology shapes how hard a controller is to optimize, and \citet{xiong2025chemotactic} study reset-free learning across two swimmer topologies. None of these works trains a population of bodies under several reset rules, which is the question we study.

\paragraph{Evaluation sensitivity.}
RL results are sensitive to implementation details~\citep{henderson2018deep, agarwal2021deep, engstrom2020implementation, andrychowicz2021what, patterson2024empirical}, and \citet{karwowski2024goodhart} document Goodhart's law in RL reward optimization. \citet{adkins2024method} make hyperparameter sensitivity a first-class axis of evaluation. We carry this axis over to embodied design, where sensitivity to termination changes the ranking of \emph{bodies}.

\section{Method}
\label{sec:method}

\subsection{Morphology Space}
\label{sec:bodies}

We use 100 articulated robots from the UNIMAL (UNIversal aniMAL) morphology space~\citep{gupta2021embodied}. DERL evolved them under three selection pressures, which define our three families:

\begin{itemize}
    \item \textbf{Floor} ($n{=}41$): evolved on flat terrain to move forward fast on predictable surfaces.
    \item \textbf{Variable terrain (VT)} ($n{=}37$): evolved on randomized terrain for robust locomotion under environmental perturbation.
    \item \textbf{Manipulation in variable terrain (MVT)} ($n{=}22$): evolved to push a box to a goal across variable terrain.
\end{itemize}

The bodies have 5 to 15 actuated hinge joints, 5 to 11 rigid parts and a kinematic tree 2 to 5 levels deep below the torso, and they differ in total mass. Every torso is a sphere of radius 0.1\,m. Every controller, including those of the MVT bodies, is trained and evaluated on the same task, walking forward on flat ground. As in DERL, the reward is the forward velocity minus a small control cost, $r_t = v_x - 0.001 \cdot \|\mathbf{a}_t\|^2$, with no alive bonus, which makes the reward of a step close to the forward speed in m/s. The controller acts every 0.02\,s. It observes the torso height and orientation, the joint angles and all velocities (21 to 41 dimensions, without the global $x$ and $y$ position or the time) and outputs hinge torques in $[-1, 1]$. Each episode starts from the body's spawn pose, with uniform noise in $[-0.01, 0.01]$ on the velocities and on every position except the global $x$ and $y$. Table~\ref{tab:env} (Appendix~\ref{sec:details_env}) lists the full setup, and Figure~\ref{fig:morph_gallery} (Appendix~\ref{sec:morph_gallery_appendix}) shows three sample bodies from each family.

\subsection{Training Protocol as an Independent Variable}
\label{sec:rules}

\begin{figure}[t]
    \centering
    \includegraphics[width=\textwidth]{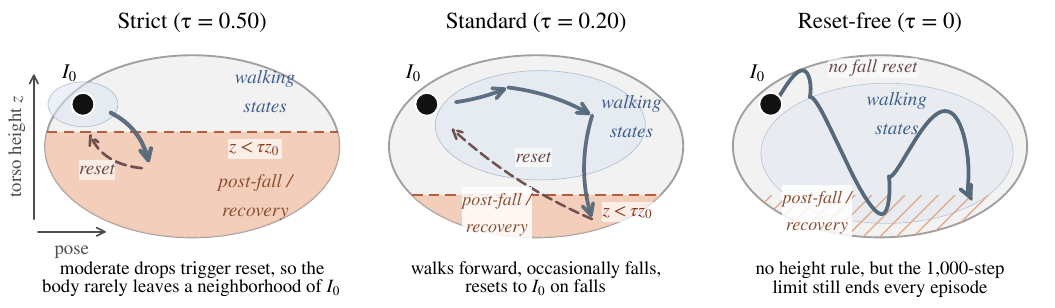}\par\vspace{2.69pt}%
    \caption{Each panel sketches the states one protocol trains on, with a generic pose dimension on the horizontal axis. $I_0$ is the start state, and reaching the region below the cutoff $z = \tau z_0$ ends an episode (Eq.~\ref{eq:termination}), except under Reset-free. \textbf{Takeaway:} Strict training keeps rollouts near standing states, while Reset-free training exposes bodies to post-fall states the controller must handle.}
    \label{fig:protocol_timeline}
\end{figure}

The reset rule has one parameter, the threshold $\tau$. During training, an episode ends early when the torso falls below the cutoff $\tau z_0$, where $z_0$ is the torso's starting height, and the body then restarts from its spawn pose. Because the reward has no alive bonus, the rule changes no step's reward, only whether an episode continues. Formally, with torso height $z_t$ and $\ell_t$ steps elapsed in the episode, the episode ends as soon as one of three conditions holds:
\begin{equation}
\label{eq:termination}
\mathrm{done}_t \;=\; \mathbf{1}\!\left[
z_t < z_{\min}(\tau)
\;\lor\;
z_t > z_{\max}
\;\lor\;
\ell_t \ge T_{\max}
\right].
\end{equation}
These are the fall cutoff, an upper height bound, and the time limit. The cutoff is $z_{\min}(\tau) = \tau z_0$ for $\tau > 0$, and Reset-free training has none, $z_{\min}(0) = -\infty$. The spawn pose holds the lowest point of the body 0.15 to 0.25\,m above the floor, so every episode starts with a short drop. The upper bound $z_{\max} = \max(3 z_0, 2\,\text{m})$ and the time limit $T_{\max} = 1000$ steps (20\,s) are the same under every rule, and the time limit ends an episode as a terminal transition, without bootstrapping from the final state (Section~\ref{sec:boundary_free} varies the time limit). Figure~\ref{fig:protocol_timeline} sketches what each rule does to the states a controller trains on. The three training protocols are:

\begin{itemize}
    \item \textbf{Standard} ($\tau = 0.20$): the reset rule of the locomotion environment our training code was adapted from (Appendix~\ref{sec:protocol_glossary_appendix}). An episode ends only after the torso drops below 20\% of its starting height.
    \item \textbf{Strict} ($\tau = 0.50$): DERL's ratio of 0.5~\citep{gupta2021embodied}, applied to $z_0$ rather than to $z_0 - 0.2$\,m. An episode ends once the torso drops below half its starting height.
    \item \textbf{Reset-free} ($\tau = 0$): without a fall cutoff, the controller keeps acting from whatever state a failed gait leaves it in. Because episodes still end at the time limit and the upper height bound, \emph{Reset-free} names the $\tau = 0$ setting only, not fully autonomous reset-free RL~\citep{sharma2022autonomous}.
\end{itemize}

Existing setups use different thresholds and do not vary them when they rank bodies. DERL~\citep{gupta2021embodied} ends an episode when the torso falls below $0.5\,(z_0 - 0.2\,\text{m})$, 0.1\,m below our Strict cutoff, and MetaMorph~\citep{gupta2022metamorph} uses the same UNIMAL environment. The default Gymnasium MuJoCo locomotion environments~\citep{towers2023gymnasium} use an \emph{absolute} height range, which corresponds to a fraction of about $0.27$ for the Ant (starting near $0.75\,\text{m}$) and 0.56 to 0.71 for Hopper, Walker2d, and Humanoid, all above Standard. No value is privileged, so we vary it.

\paragraph{Per-body cutoff check.}\label{sec:cutoff_check} For some bodies, a rule does not act as its value suggests (Appendix~\ref{sec:fallrules}). A torso resting on the floor sits at about its radius, 0.1\,m. For the 28 shortest bodies, the Standard cutoff lies below that, so a resting torso stays above it, and Standard training is in effect Reset-free training. The Strict cutoff has the opposite problem: it can end an episode during the drop from the spawn pose. With zero action, 45 of the 100 bodies are below it 1\,s after leaving the spawn pose, so a Strict episode continues only if the controller catches the drop, which most controllers learn to do. For 11 bodies, however, the Strict-trained runs end within 20 steps (0.4\,s) on average when evaluated with the Strict cutoff. We call these the cut-off bodies. Ten of them are among the 45, and 7 are among the 28 shortest. We therefore also report the main results on the 72 taller bodies and without the 11 cut-off bodies, as post hoc checks (Appendix~\ref{sec:ranking_appendix}).

\subsection{Experimental Design}
\label{sec:experimental_design}

\paragraph{Main sweep.}
We train 900 PPO~\citep{schulman2017proximal} controllers, one for each combination of body, protocol and seed ($100 \times 3 \times 3$), each from its own random initialization. Each controller trains for $5 \times 10^6$ environment steps in MuJoCo~\citep{todorov2012mujoco}, simulated on GPUs with MuJoCo MJX~\citep{mjx2024}. All runs are specified with the same PPO hyperparameters (Appendix~\ref{sec:appendix_hyperparams}, Table~\ref{tab:hyperparams}), so no protocol receives its own tuning.

We evaluate each controller on 20 episodes. So that controllers trained under different protocols are compared on equal terms, the main results use one \emph{common evaluator}, which ends episodes at the Standard cutoff ($\tau_{\text{eval}} = 0.20$). Because this evaluator shares its cutoff with Standard training, Section~\ref{sec:distance_audit} checks the main result with a \emph{neutral evaluator} that has no cutoff ($\tau_{\text{eval}} = 0$) but keeps the time limit. A Strict evaluator ($\tau_{\text{eval}} = 0.50$) identifies the cut-off runs of Section~\ref{sec:cutoff_check}.

\paragraph{Cross-algorithm validation.}
To check that the contrast between Strict and Reset-free training does not depend on PPO, we also train SAC~\citep{haarnoja2018soft}, an off-policy algorithm, with Brax's implementation~\citep{freeman2021brax} in the same MJX environment. We train it on 24 bodies (6 Floor, 13 VT, 5 MVT) under all three protocols with three seeds and $5 \times 10^6$ environment steps each ($216$ runs). We chose these bodies to cover the range of outcomes under Strict training, and Section~\ref{sec:cross_algo} and Appendix~\ref{sec:sac_appendix} give the results.

\subsection{Statistical Analysis}
\label{sec:stats}

We score each controller by its \emph{per-step reward} (PSR), the reward per step of each evaluation episode, averaged over its 20 episodes. A body's score under a protocol is the mean over its three seeds. Because this average covers only the steps before an episode ends, Appendix~\ref{sec:per_seed_metric} repeats the family comparison with the episode return divided by $T_{\max}$, which counts the steps after an early end as zero reward, and finds the same difference. A body's \emph{retention} after Reset-free or Strict training is the share of its Standard reward that it keeps (RF/Std or Str/Std), and a family's \emph{loss} is one minus the mean retention of its bodies. Total distance traveled is a secondary metric.

Our main statistical question is whether the protocol affects the three families differently. The three seeds of a body are repeated measurements of that body, so the body is the unit of analysis. The primary test is a permutation test that shuffles family labels across bodies and recomputes the protocol-by-family $F$ statistic, checked by a Wald test with standard errors clustered by body. An ANOVA on all 900 runs, which treats seeds as independent, is only a supplement (Appendix~\ref{sec:per_seed_appendix} adds seed-averaged, rank-transformed, mixed-effects and generalized estimating equation (GEE) variants). We call a test planned if its decision rule was fixed before its outcome was known.

A rank correlation is hard to judge on its own, because even two runs of one protocol with different seeds rank the bodies differently. We therefore compare each correlation between two protocols with a \emph{seed-noise reference}: the correlation we would expect if seed noise were the only difference between them. We estimate it from how well the single-seed rankings within each protocol agree, adjusted for averaging over three seeds with the Spearman--Brown formula (Appendix~\ref{sec:ranking_appendix}). Confidence intervals for correlations and selection results are percentile bootstraps over $B = 10{,}000$ resamples of the bodies. The seed-noise references and the analyses of body subsets are post hoc.

\section{Results}
\label{sec:results}

Changing the reset rule during training changes how much each family loses (Section~\ref{sec:familycost}) and which bodies rank highest (Section~\ref{sec:reorder}), yet ranking bodies by their worst protocol still keeps most of the reward of each protocol's own best bodies (Section~\ref{sec:regret}). Unless stated otherwise, results are per-step reward under the common evaluator ($\tau_{\text{eval}} = 0.20$), and the main results also hold without the bodies whose cutoffs misfire (Section~\ref{sec:cutoff_check}, Appendix~\ref{sec:ranking_appendix}).

\subsection{Reset-Free Training Costs Floor Bodies Most}
\label{sec:familycost}

\begin{table}[t]
\centering
\caption{Family means of per-step reward over $n$ bodies (seed-averaged, common evaluator). RF/Std and Str/Std average each body's share of its Standard reward, so they are not ratios of the family means. Strict values include the cut-off runs, which lower VT's Str/Std (Section~\ref{sec:cutoff_check}). \textbf{Takeaway:} without fall resets, Floor bodies keep far less of their reward than VT bodies.}
\label{tab:family}
\small
\begin{tabular}{lcrrrcc}
\toprule
Family & $n$ & Standard $\uparrow$ & Strict $\uparrow$ & Reset-free $\uparrow$ & RF/Std $\uparrow$ & Str/Std $\uparrow$ \\
\midrule
Floor & 41 & 2.428 & 1.837 & 1.453 & 0.55 & 0.78 \\
VT    & 37 & 2.304 & 1.467 & 1.999 & 0.84 & 0.66 \\
MVT   & 22 & 2.651 & 2.317 & 2.088 & 0.72 & 0.91 \\
\bottomrule
\end{tabular}
\end{table}

\begin{figure}[t]
    \centering
    \includegraphics[width=\textwidth]{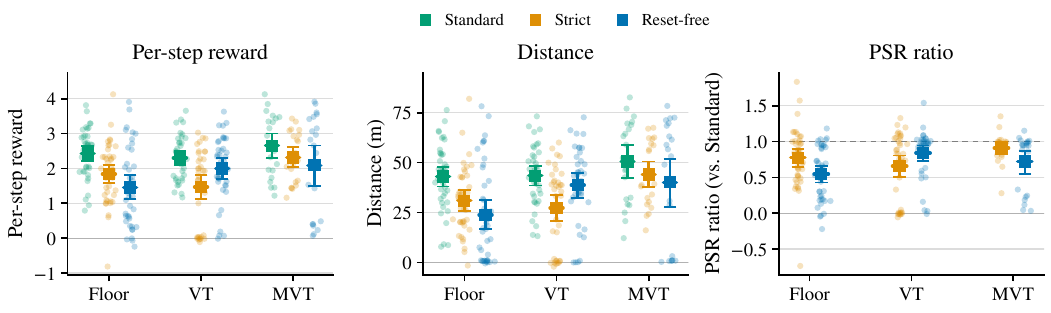}\par\vspace{2.36pt}%
    \caption{Per-step reward (left), distance (middle), and retention, the reward relative to Standard training (right) for each family. Points are bodies, and large markers with intervals show family means with 95\% bootstrap CIs over bodies. Strict values near zero come from cut-off runs (Section~\ref{sec:cutoff_check}). \textbf{Takeaway:} without fall resets, the whole Floor distribution shifts down relative to VT, not just a few bodies.}
    \label{fig:hero}
\end{figure}

Removing fall resets costs Floor bodies far more than VT bodies (Table~\ref{tab:family}). After Reset-free training, Floor bodies lose on average 45\% of the reward they reach after Standard training, VT bodies 16\% and MVT bodies 28\%. Every family loses more on the 72 taller bodies, where the Standard cutoff can act, and Floor still loses the most. This is consistent with the reset rule acting through the states a controller trains on: without a cutoff, a fall no longer ends the episode, so the states that follow it stay in training. Distance traveled shows the same pattern (Figure~\ref{fig:hero}, middle, and Appendix~\ref{sec:locomotion_quality_appendix}).

\subsection{Protocol Choice Changes Which Bodies Appear Optimal}
\label{sec:reorder}

\begin{figure}[t]
    \centering
    \includegraphics[width=\textwidth]{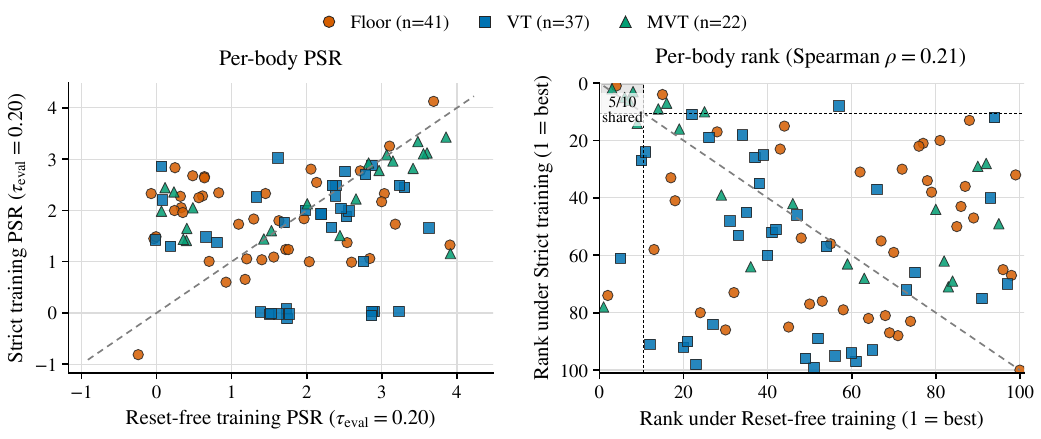}\par\vspace{0.80pt}%
    \caption{Reset-free vs.\ Strict training per body (common evaluator). Left: per-step reward. Floor bodies lie mostly above the diagonal, where Strict training does better, and VT bodies mostly below it. Strict values near zero come from cut-off runs (Section~\ref{sec:cutoff_check}). Right: ranks, with dotted lines at the top-10 cutoff. \textbf{Takeaway:} Reset-free and Strict training rank the bodies very differently, even within the Floor and VT families ($\rho = 0.02$ and $0.13$).}
    \label{fig:rf_vs_strict}
\end{figure}

\begin{figure}[t]
    \centering
    \includegraphics[width=\textwidth]{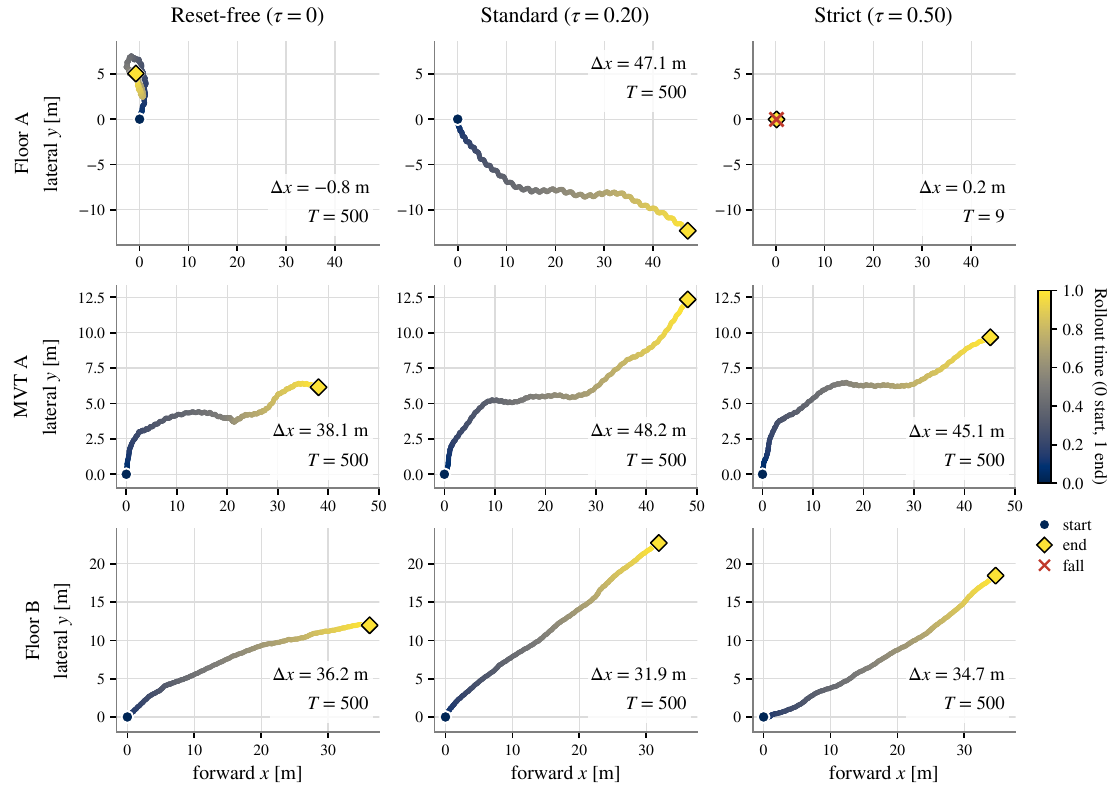}\par\vspace{3.02pt}%
    \caption{Top-down torso $(x,y)$ trajectories from illustrative retrains of three bodies under three training protocols. Panels without early termination show the first 500 control steps (10\,s). \emph{Floor A} walks under Standard and makes no forward progress under Reset-free, and its Strict episode ends at step~9. \emph{MVT A} and \emph{Floor B} walk under all three. Full IDs are in Appendix~\ref{sec:repro_appendix} and chronophotographs in Figures~\ref{fig:gait_chronophoto_hero} and~\ref{fig:gait_chronophoto_full}. \textbf{Takeaway:} controllers of the same body can differ in their whole path, not only in the distance they cover.}
    \label{fig:gait_temporal}
\end{figure}

The protocol also changes which bodies rank highest (Figure~\ref{fig:rf_vs_strict}). After Reset-free and Strict training, the rankings of all 100 bodies correlate at $\rho = 0.21$. If seed noise were the only difference between the two protocols, they would agree at about $0.89$ (Table~\ref{tab:rankings}). The gap persists on the 72 taller bodies ($\rho = 0.26$ against a reference of $0.89$) and without the 11 cut-off bodies ($0.25$ against $0.85$). The top-20 lists of Reset-free and Strict training share only 10 bodies, against about 16 in simulated three-seed reruns of one protocol (Appendix~\ref{sec:ranking_appendix}). Every pair of protocols agrees less than its seed-noise reference. Measured on single-seed rankings, the excess disagreement, which is zero when changing the rule moves a ranking no more than a new seed does, is 0.53 [0.36, 0.70] for Reset-free versus Strict, 0.29 [0.17, 0.43] for Standard versus Strict and 0.17 [0.08, 0.28] for Reset-free versus Standard (post hoc, Appendix~\ref{sec:ranking_appendix}). Standard and Reset-free agree the most ($\rho = 0.72$), partly because their rules coincide for the 28 short bodies. Scored without an evaluation cutoff, the Reset-free and Strict rankings still correlate at only $0.31$. A control agrees far more: on 60 bodies that never fell below a tenth of their starting height in our probes, retraining with a 0.10 threshold ranks them nearly as Reset-free training does ($\rho = 0.92$), while Reset-free and Strict training of the same bodies give $\rho = 0.44$ (post hoc, Appendix~\ref{sec:ranking_appendix}). Figure~\ref{fig:gait_temporal} and Appendix~\ref{sec:chronophoto_appendix} show single bodies under each protocol.

\begin{table}[t]
\centering
\caption{Spearman correlations between the protocols' rankings of the 100 bodies by per-step reward (mean over three seeds, common evaluator), with 95\% percentile bootstrap CIs over bodies ($B = 10{,}000$). The seed-noise reference is the correlation expected if seed noise were the only difference between the two protocols (Section~\ref{sec:stats}). \textbf{Takeaway:} every pair of protocols ranks the bodies less alike than seed noise alone would, most of all Reset-free and Strict.}
\label{tab:rankings}
\small
\begin{tabular}{lcc}
\toprule
Comparison & Spearman $\rho$ [95\% CI] & Seed-noise reference [95\% CI] \\
\midrule
Standard vs.\ Reset-free & 0.72 [0.58, 0.83] & 0.91 [0.87, 0.94] \\
Standard vs.\ Strict     & 0.46 [0.26, 0.62] & 0.87 [0.82, 0.91] \\
Reset-free vs.\ Strict   & 0.21 [$-0.02$, 0.40] & 0.89 [0.84, 0.92] \\
\bottomrule
\end{tabular}
\end{table}

\subsection{Protocol-Robust Selection}
\label{sec:regret}

\begin{table}[t]
\centering
\caption{Worst-case retention by criterion. Each row picks $k$ bodies by one criterion. For each protocol, we divide the mean reward of the picked bodies by that of the protocol's own top $k$, and each cell reports the smallest of the three ratios (in sample, common evaluator). $^\dagger$ marks the selector we use, and Appendix~\ref{sec:robust_selection_appendix} reports each protocol separately with bootstrap CIs. \textbf{Takeaway:} choosing bodies with all three protocols in view avoids the large losses of choosing by one protocol.}
\label{tab:robust_selection}
\small
\begin{tabular}{lccc}
\toprule
Selection criterion & $k{=}5 \uparrow$ & $k{=}10 \uparrow$ & $k{=}20 \uparrow$ \\
\midrule
Standard-only                          & 49.0\% & 74.9\% & 85.1\% \\
Strict-only                            & 91.1\% & 87.9\% & 77.4\% \\
Reset-free-only                        & 67.7\% & 81.9\% & 78.6\% \\
Mean over protocols                    & 92.3\% & 93.9\% & 89.9\% \\
Min over protocols$^\dagger$ (robust)  & 92.3\% & 92.8\% & 91.4\% \\
\bottomrule
\end{tabular}
\end{table}

When the reset rule is not known in advance, the useful bodies are those that stay strong under every plausible protocol. A simple way to find them is to score each body by its worst reward across the three protocols and keep the top $k$. In sample, these bodies keep at least 91\% of the mean reward of each protocol's own top $k$, for $k = 5$, 10 and 20 (Table~\ref{tab:robust_selection}), and 83 to 85\% averaged across held-out seeds, the most of the five criteria (post hoc, Appendix~\ref{sec:robust_selection_appendix}). Choosing by one protocol is riskier: the five best bodies after Standard training, once trained under Strict, earn only $49.0\%$ of the reward of the five best Strict-trained bodies ($75.7\%$ on the 72 taller bodies). The selector describes performance across the tested protocols, and a planned test with one protocol held out did not support it as a predictor for an untested protocol (Appendix~\ref{sec:robust_selection_appendix}).

\subsection{Statistical Tests of the Family Difference}
\label{sec:interaction}

The permutation test finds that the three families respond differently to the three protocols ($p = 0.004$), and the Wald test with body-clustered standard errors ($p = 0.0014$) and the supplementary ANOVA agree (Appendix~\ref{sec:per_seed_alt_models}). A mixed-effects model with a random intercept per body gives a VT-by-Reset-free interaction of 0.67 [0.39, 0.95] (Table~\ref{tab:interaction_robustness}). On the Floor and VT bodies alone, the interaction also holds in a body-level permutation test ($p < 0.01$, Appendix~\ref{sec:per_seed_lofo}). Under Strict training, the family difference depends on the 11 cut-off bodies (Table~\ref{tab:survive}). The central contrast, that Floor bodies lose more than VT bodies without fall resets, holds on every body set: the share of Standard reward that Floor bodies keep after Reset-free training lies 0.30 below VT's on all 100 bodies, 0.32 below on the 72 taller bodies and 0.25 below without the 11 cut-off bodies (post hoc permutation $p < 0.01$, Table~\ref{tab:survive}). It also holds when both problems are removed at once, on the 65 of the 72 taller bodies with no cut-off Strict run (Appendix~\ref{sec:ranking_appendix}), in four further robustness tests (Table~\ref{tab:interaction_robustness}) and in all 1,000 subsamples with 22 bodies per family (Appendix~\ref{sec:per_seed_balanced}).

\subsection{Cross-Algorithm Validation with SAC}
\label{sec:cross_algo}

SAC orders the families as PPO does. The SAC runs were set up to check the contrast between Strict and Reset-free training, so here we use the ratio of a body's Strict to its Reset-free reward, not RF/Std as in Section~\ref{sec:familycost}. On these 24 bodies, chosen to cover the range of outcomes under Strict training (Section~\ref{sec:experimental_design}), both algorithms give Floor bodies the highest mean ratio and MVT bodies the lowest (2.18, 1.27 and 0.91 for Floor, VT and MVT under PPO, and 2.41, 1.52 and 0.96 under SAC, each over its bodies whose Reset-free reward exceeds 0.05). The two algorithms also agree body by body: their ratios correlate at Spearman $\rho = 0.75$ on the 23 bodies where this holds under both, and at 0.72 to 0.83 with any one body left out. SAC thus supports the direction of the family effect (Appendix~\ref{sec:sac_appendix}).

\subsection{Robustness to the Evaluator and a Distance Audit}
\label{sec:distance_audit}

The family difference also holds when runs are scored without a cutoff. Under the neutral evaluator, Floor bodies still keep the least of their Standard reward after Reset-free training: 64\%, against 95\% for VT and 85\% for MVT (Appendix~\ref{sec:regret_robustness_appendix}). Distance points the same way. Without a cutoff at evaluation, Reset-free-trained Floor controllers cover $31.2$\,m per episode on average, against $44.5$\,m after Standard training (Appendix~\ref{sec:compute_budget_appendix}). Appendix~\ref{sec:locomotion_quality_appendix} shows that the reward gap reflects slower gaits rather than controllers that stop moving, and Appendix~\ref{sec:pareto_appendix} that control effort is similar across protocols.

\subsection{Replication Across Seeds and Body Structure}
\label{sec:baseline_control}

The family difference replicates in each of the three seeds, with the same ordering of Reset-free retention: Floor bodies keep 53 to 59\% of their Standard reward, against 83 to 88\% for VT bodies (Appendix~\ref{sec:per_seed_appendix}). Here a body's sensitivity is the gap, not the ratio, between its Strict and Reset-free reward. In our corpus, five simple descriptors of body structure, such as starting height and limb count, explain 46\% of the variation in this gap, while a body's overall skill, its Standard reward, explains 11\%. Taller bodies and bodies with deeper limb trees also keep less of their reward after Reset-free training (Spearman $\rho = -0.53$ and $-0.48$ with RF/Std), and bodies with more first-level limbs keep more ($\rho = 0.50$). With starting height and limb count controlled, the families still differ in this gap (Appendix~\ref{sec:features_appendix}), so these two descriptors do not account for the family difference.

\subsection{Time Limit and Other Protocol Variables}
\label{sec:boundary_free}

Our Reset-free training removes the fall cutoff but keeps the 1,000-step time limit, so our results describe training without fall resets, not fully autonomous training. A planned experiment retrained 30 bodies under Reset-free with no time limit at all. Their ranking correlates with the ordinary Reset-free ranking at $\rho = 0.85$ (95\% CI $[0.67,0.93]$). This meets the plan's criterion for a changed ranking (interval below 0.95), and the Floor and MVT family means drop (Figure~\ref{fig:boundary_free_forest}). These runs, however, used a later version of our training code, which by itself lowers Reset-free Floor scores by a similar amount (post hoc), so the experiment does not isolate the time limit. On the same bodies, a planned factorial test finds that the reset rule and the time limit interact ($p = 0.025$), and Reset-free rankings with limits of 3,001 and 10,001 steps correlate with the ordinary ones at 0.92 and 0.85 (Appendix~\ref{sec:boundary_scale_appendix}). Appendix~\ref{sec:otherspaces_appendix} applies the same ranking test to other protocol variables in Ant, Evolution Gym and Humanoid bodies.

\section{Discussion}
\label{sec:discussion}

\paragraph{Morphology modulates sensitivity to the reset rule.}
For co-design, a body's score under one reset rule mixes how good the body is with how well it suits that rule, and optimizing it selects for both. Across many bodies, the cost of removing fall resets is a property of the body. Floor bodies lose about three times as much of their reward as VT bodies, which fits the evolutionary account of Section~\ref{sec:intro}. Reset-free RL usually studies one fixed embodiment~\citep{sharma2022autonomous, yang2025morefree, gupta2021mtrf, eysenbach2018leave}, and our results suggest testing reset-free RL methods on several body families, with Floor bodies, which lose the most, as stress tests for recovery from falls (Appendix~\ref{sec:extended_discussion_appendix}). Universal controllers trained across many bodies, such as MetaMorph~\citep{gupta2022metamorph} and URMA~\citep{bohlinger2024onepolicy}, are a natural next test of whether one policy can absorb this interaction.

\paragraph{A pragmatic audit.}
\label{sec:audit}
A ranking from a single training protocol can be specific to that protocol, so a co-design benchmark should test protocol sensitivity before it names a winning body. The test need not be a full sweep. Retrain the top $k$ bodies, or a subset stratified by family, under one looser and one stricter threshold (120 runs for 20 bodies and three seeds), score every controller with a common evaluator, and report:

\vspace{-0.35em}
\begin{table}[H]
\centering
\footnotesize
\setlength{\tabcolsep}{3pt}
\begin{tabular}{@{}p{0.14\linewidth}p{0.38\linewidth}p{0.44\linewidth}@{}}
\toprule
Field & Report & This paper \\
\midrule
Training reset & Predicate, thresholds, other mechanics & $\tau_{\text{train}}\!\in\!\{0,0.20,0.50\}$; all else specified identically \\
Cutoff check & Each body's cutoff vs.\ torso and reset drop & 28 Standard cutoffs below the torso, 11 cut-off bodies \\
Evaluator & Common reset and metric & $\tau_{\text{eval}} = 0.20$; per-step reward \\
Audit set & Bodies, seeds, settings & 100 bodies $\times$ 3 protocols $\times$ 3 seeds; $k = 5, 10, 20$ \\
Statistics & Rank agreement against seeds, retention, CIs & Spearman $\rho$ vs.\ seeds; top-10 overlap; top-$k$ retention \\
Selector & Rule for declaring a best body & Minimum reward across protocols (in sample) \\
\bottomrule
\end{tabular}
\end{table}
\vspace{-0.55em}

If the top bodies change by more than they do between seeds, the benchmark has found a protocol-specific ranking. The seed-noise reference needs no extra runs, because it comes from the agreement between single-seed rankings within each protocol, and retraining with a threshold below the torso's resting height, as in Section~\ref{sec:reorder}, gives a control for differences between separate training runs. The per-body cutoff check is worth running on standard setups too. DERL's own rule puts the cutoff of our 25 bodies that start at 0.3\,m at half the torso radius, and the lower height bound of Gymnasium's Ant, 0.2\,m, lies below its own 0.25\,m torso radius, so in both cases a torso resting on the floor stays above the cutoff (Appendix~\ref{sec:fallrules_rationale}).

\paragraph{Released artifacts.}
\label{sec:artifacts}
Everything is in the public repository at \codeurl{}: the training and analysis code, the run summaries of the PPO and SAC UNIMAL experiments and of 5,860 runs on Humanoid, Ant and Evolution Gym bodies, and a seeded CPU script that recomputes 2,401 values from them in a few minutes without GPU training. The code is under the MIT license and the run records under CC BY 4.0, with a data card and Croissant metadata (Appendix~\ref{sec:records}, Table~\ref{tab:artifacts}).

\paragraph{Scope and limitations.}
\label{sec:limitations}
This study establishes that changing the fall-termination threshold reorders DERL-evolved UNIMAL morphologies and produces a family-specific interaction in this corpus. The named family ordering is not claimed for other design spaces. More generally, which morphology ranks best can depend on the controller-training protocol used to score it, so that protocol should be reported and reasonable alternatives audited. Whether a particular benchmark is sensitive remains an empirical question. Our experiments use height-based resets, a fixed compute budget, and simulated UNIMAL bodies trained on a single locomotion task. SAC supports the direction of the family ordering (Section~\ref{sec:cross_algo}), but loss percentages are PPO estimates, and tuning the learner for each protocol or moving to hardware or other morphology spaces could change them. The family difference also need not hold for other termination rules. In a planned test with a rule ending episodes when the torso touched the ground, a body's sensitivity to that rule correlated only weakly with its sensitivity to the height cutoff ($\rho = 0.19$, below the planned 0.60).

\section{Conclusion}
\label{sec:conclusion}

Which morphology ranks best can depend on the controller-training protocol used to score it. In the UNIMAL corpus, changing the fall threshold, with the other settings specified identically, changes rank agreement beyond seed variation, the top-ranked bodies, and the response of the three evolutionary families to Reset-free training. The specific family ordering remains scoped to this corpus. Reporting the reset rule, checking that it acts on every body, and using a common evaluator are low-cost specification safeguards that do not assume every benchmark will be sensitive. Benchmark authors should then test whether selected bodies survive reasonable protocol alternatives and distinguish setting-specific from protocol-robust selections. Because real robots often cannot rely on cheap fall resets, co-design methods that select bodies for hardware should treat the training protocol as part of the design problem.

\begin{ack}
This work was partially supported by the National Science Centre, Poland, under PRELUDIUM Grant No.~2024/53/N/ST6/03370 and Sonata Bis Grant No.~2024/54/E/ST6/00388. We gratefully acknowledge Polish high-performance computing infrastructure PLGrid (HPC Centers: ACK Cyfronet AGH) for providing computer facilities and support within computational grant no.~PLG/2026/019509. We thank the Faculty of Mathematics, Informatics and Mechanics of the University of Warsaw for access to its Entropy GPU cluster.

\end{ack}

\bibliography{references}
\bibliographystyle{plainnat}

\clearpage
\appendix
\raggedbottom
\begin{center}
{\Large\bfseries Appendix Contents}
\end{center}
\vspace{-0.2em}
\noindent\rule{\textwidth}{0.4pt}
\vspace{-0.4em}
\begingroup

\apptocmain{A~~Experimental Details}{sec:appendix}
\apptocsub{A.1~~Training Hyperparameters}{sec:appendix_hyperparams}
\apptocsub{A.2~~Environment Details}{sec:appendix_env}
\apptocsub{A.3~~Morphology Space}{sec:appendix_morph}
\apptocsub{A.4~~Compute}{sec:compute}
\apptocsub{A.5~~Data Validation}{sec:appendix_validation}

\apptocmain{B~~Detailed Reproducibility Analysis}{sec:repro_appendix}
\apptocsub{B.1~~Released Artifacts}{sec:records}

\apptocmain{C~~Protocol Glossary and Per-Body Cutoff Checks}{sec:protocol_glossary_appendix}
\apptocsub{C.1~~Per-Body Cutoff Checks}{sec:fallrules}

\apptocmain{D~~Evaluation-Protocol Robustness}{sec:regret_robustness_appendix}

\apptocmain{E~~Protocol-Robust Selection: Full Retention Matrix}{sec:robust_selection_appendix}

\apptocmain{F~~Continuous-Threshold Sweep}{sec:threshold_landscape_appendix}

\apptocmain{G~~Within-Family Ratio Distributions}{sec:ratio_distributions_appendix}

\apptocmain{H~~Per-Seed Replication and Alternative Blocked Models}{sec:per_seed_appendix}
\apptocsub{H.1~~Robustness of the Protocol-by-Family Interaction}{sec:per_seed_robustness}
\apptocsub{H.2~~Balanced-Family Robustness}{sec:per_seed_balanced}
\apptocsub{H.3~~Leave-One-Family-Out Robustness}{sec:per_seed_lofo}
\apptocsub{H.4~~Per-Seed RF/Std ANOVA}{sec:per_seed_rf_std}
\apptocsub{H.5~~Alternative Blocked Models}{sec:per_seed_alt_models}
\apptocsub{H.6~~Morphology-Level Permutation Test}{sec:per_seed_permutation}
\apptocsub{H.7~~Metric Sensitivity}{sec:per_seed_metric}
\apptocsub{H.8~~Compute-Budget Audit}{sec:compute_budget_appendix}

\apptocmain{I~~Morphological Features Track Sensitivity}{sec:features_appendix}

\apptocmain{J~~Cross-Algorithm Validation with SAC}{sec:sac_appendix}

\apptocmain{K~~Morphology Gallery}{sec:morph_gallery_appendix}

\apptocmain{L~~Rank Agreement Across Protocols and Seeds}{sec:rf_vs_strict_appendix}
\apptocsub{L.1~~Main Results on Each Body Set and the Seed-Noise Reference}{sec:ranking_appendix}

\apptocmain{M~~Locomotion-Quality Audit}{sec:locomotion_quality_appendix}

\apptocmain{N~~Control-Effort Confound Audit}{sec:pareto_appendix}

\apptocmain{O~~Extended Discussion}{sec:extended_discussion_appendix}

\apptocmain{P~~Time Limit and Other Protocol Variables}{sec:boundary_scale_appendix}
\apptocsub{P.1~~Other Design Spaces and Protocol Variables}{sec:otherspaces_appendix}

\apptocmain{Q~~Companion Chronophotograph}{sec:chronophoto_appendix}

\endgroup
\clearpage

\section{Experimental Details}
\label{sec:appendix}

\subsection{Training Hyperparameters}
\label{sec:appendix_hyperparams}

All PPO training is specified with identical hyperparameters across thresholds, morphologies, and seeds. The run records store only the rule, seed, and step budget, so the other entries of Table~\ref{tab:hyperparams} are the values set by the launch code.

\begin{table}[h]
\centering
\caption{PPO training hyperparameters for the MJX main sweep.}
\label{tab:hyperparams}
\begin{tabular}{lr}
\toprule
Parameter & Value \\
\midrule
Algorithm & PPO~\citep{schulman2017proximal} \\
Implementation & Custom PPO in JAX on MuJoCo MJX \\
Total environment steps & 5,000,000 \\
Parallel environments & 128 \\
Steps per rollout & 16 \\
Minibatch size & 256 \\
PPO epochs per update & 10 \\
Learning rate & $3 \times 10^{-4}$ \\
Discount ($\gamma$) & 0.99 \\
GAE ($\lambda$) & 0.95 \\
Clip range ($\epsilon$) & 0.2 \\
Value loss coefficient & 0.5 \\
Entropy coefficient & 0.01, applied to half the policy entropy \\
Maximum gradient norm & 0.5 \\
Optimizer & Adam \\
Advantage normalization & Per minibatch \\
Policy network & MLP [64, 64] \\
Value network & MLP [64, 64] \\
Activation & Tanh \\
Max episode length ($T_{\max}$) & 1,000 steps \\
Physics timestep & 0.005s \\
Frame skip & 4 \\
Effective control dt & 0.020s \\
\bottomrule
\end{tabular}
\end{table}

\subsection{Environment Details}
\label{sec:appendix_env}
\label{sec:details_env}

\begin{table}[htbp]
\centering
\caption{Training and evaluation setup. Every setting is specified identically for the three fall rules except the threshold $\tau$, and the last column notes a public setup with the same value where one exists.}
\label{tab:env}
\footnotesize
\setlength{\tabcolsep}{4pt}
\begin{tabularx}{\linewidth}{@{}L{0.14\linewidth}YL{0.17\linewidth}@{}}
\toprule
Setting & Value & Public reference \\
\midrule
Bodies & 100 UNIMAL bodies (41 Floor, 37 VT, 22 MVT, 97 distinct files), torso a sphere of radius 0.1\,m & UNIMAL-100 benchmark (DERL-evolved) \\
Task, reward & Forward locomotion on flat ground, $r_t = v_x - 0.001\,\lVert a_t \rVert^2$, no alive bonus & DERL also has no alive bonus \\
Observations, actions & Torso height and orientation, joint angles and all velocities (21 to 41 dimensions, no global $x$, $y$ or time). Hinge torques in $[-1, 1]$ (5 to 15 dimensions) & \\
Reset & Spawn pose (torso at reset height $z_0$, lowest point 0.15 to 0.25\,m above the floor), uniform noise in $[-0.01, 0.01]$ on velocities and on positions except global $x$, $y$ & Spawn heights from DERL files \\
Episode end & Fall ($z_t < \tau z_0$, none at $\tau = 0$), upper bound $z_t > \max(3 z_0, 2\,\text{m})$, or 1,000 steps, a terminal transition (not bootstrapped) under every rule & DERL also uses 1,000 steps but bootstraps \\
Fall rule & $\tau \in \{0, 0.20, 0.50\}$, cutoffs 0.06 to 0.21\,m (Standard) and 0.15 to 0.52\,m (Strict), a later sweep adds 0.10 and 0.35 & Section~\ref{sec:rules} \\
Training & PPO, one preset for every rule (Appendix~\ref{sec:appendix}), 5M environment steps, seeds 42 to 44, MuJoCo MJX & 5M steps, DERL's per-body budget \\
Evaluation & 20 evaluation episodes under the neutral (no fall clause), common ($\tau_{\text{eval}} = 0.20$) and Strict evaluators, scored by per-step reward (PSR) and fixed-horizon return (episode return divided by 1,000) & Section~\ref{sec:stats} \\
\bottomrule
\end{tabularx}
\end{table}

\paragraph{Simulator.} MuJoCo 3.x~\citep{todorov2012mujoco} with MJX acceleration for training.

\paragraph{Reward function.} $r_t = v_x - 0.001 \cdot \|\mathbf{a}_t\|^2$, where $v_x = (x_{t+1} - x_t) / \Delta t$ is the forward velocity and $\mathbf{a}_t \in [-1, 1]^{n_{\text{act}}}$ is the normalized action vector.

\paragraph{Termination.} As formalized in Equation~\ref{eq:termination}, the episode ends on a fall ($z_t < \tau z_0$ when $\tau > 0$), on the upper height bound ($z_t > z_{\max} = \max(3 z_0, 2\,\text{m})$), or at the shared horizon $T_{\max} = 1000$, which is a terminal transition (not bootstrapped) under every rule. Here $z_0$ is the torso height in the reset (spawn) pose, which places the lowest point of the body 0.15 to 0.25\,m above the floor, so every episode begins with a short drop. At $\tau = 0$ the fall clause is inactive; the shared horizon and upper height bound still apply.

\paragraph{Evaluation.} Each trained policy is evaluated for 20 episodes under three evaluation settings: $\tau_{\text{eval}} = 0.20$, $\tau_{\text{eval}} = 0.50$, and neutral evaluation ($\tau_{\text{eval}} = 0$). The main text uses $\tau_{\text{eval}} = 0.20$ unless noted otherwise. For PPO, actions are sampled stochastically (from the Gaussian policy distribution) at evaluation to match the training-time action distribution across all conditions. Rank comparisons use PSR averaged over evaluation episodes and training seeds (Table~\ref{tab:env}).

\subsection{Morphology Space}
\label{sec:appendix_morph}

We use the 100 UNIMAL morphologies of MetaMorph's UNIMAL-100 training set~\citep{gupta2022metamorph}, evolved by DERL~\citep{gupta2021embodied}, comprising three evolutionary families:

\begin{itemize}
    \item \textbf{Floor} ($n=41$): Evolved for flat-terrain locomotion.
    \item \textbf{VT} ($n=37$): Evolved for variable-terrain locomotion (slopes, steps, rugged ground).
    \item \textbf{MVT} ($n=22$): Evolved for manipulation in variable terrain (pushing a box to a goal).
\end{itemize}

Morphologies vary in number of rigid bodies (5--11), actuated hinge joints (5--15), total mass, kinematic tree depth (2--5 levels below the torso), and limb geometry. All share a central torso with tree-structured limbs.

Three pairs of the 100 body files are byte-identical, and dropping one body of each pair changes no three-seed rank correlation between rules by more than 0.04.

\Needspace{0.42\textheight}%
\subsection{Compute}
\label{sec:compute}

All UNIMAL training ran on a shared GPU cluster, with up to four PPO processes packed per GPU. The corpus records do not store the GPU model, and the panels that recorded it ran on NVIDIA TITAN V (12~GB), GeForce RTX 2080 Ti, TITAN X (Pascal), and GeForce GTX 1080 Ti GPUs. Typical wall time is $1.09$ hours per 5M-step PPO run of the corpus, which ran one process per GPU. The table below sums the recorded run time of each released panel's runs (for the UNIMAL panels, training plus evaluation). It equals GPU-hours where each run had its own GPU, and where several runs shared a GPU, GPU-hours are lower: the 108 runs of the retraining with later code took $228$ hours of wall time at four runs per GPU, about $57$ GPU-hours. The Humanoid and Ant panels ran on NVIDIA GH200 and H100 GPUs, and the Evolution Gym panel on CPUs.

\begin{center}
\footnotesize
\begin{tabular}{@{}llrr@{}}
\toprule
Panel & Hardware & Runs & Run-hours \\
\midrule
Main corpus (Section~\ref{sec:method}) & GPU, 1 run per GPU & 900 & 980 \\
Thresholds 0.10 and 0.35 (Appendix~\ref{sec:threshold_landscape_appendix}) & GPU, 4 runs per GPU & 600 & 1,006 \\
SAC panel (Appendix~\ref{sec:sac_appendix}) & GPU, 1 run per GPU & 216 & 105 \\
Retraining with later code & GPU, 4 runs per GPU & 108 & 228 \\
Structural transfer to non-evolved bodies & GPU, up to 4 per GPU & 540 & 1,025 \\
Existing robots & GPU, up to 4 per GPU & 60 & 53 \\
Longer budget & GPU, up to 4 per GPU & 162 & 183 \\
Contact-based termination & GPU, up to 4 per GPU & 270 & 552 \\
No time limit & GPU, up to 4 per GPU & 90 & 180 \\
Rule by time limit & GPU, up to 4 per GPU & 180 & 370 \\
Time-limit durations & GPU, up to 4 per GPU & 180 & 357 \\
Humanoid, fall rules (Appendix~\ref{sec:otherspaces_appendix}) & NVIDIA GH200 & 1,800 & 1,427 \\
Humanoid, time limit & NVIDIA GH200 & 900 & 725 \\
Ant, reset initialization & NVIDIA H100 & 160 & 85 \\
Ant, three other variables & NVIDIA GH200 & 1,800 & not recorded \\
Evolution Gym, time limit & CPU & 1,200 & 2,155 \\
\bottomrule
\end{tabular}
\end{center}

\subsection{Data Validation}
\label{sec:appendix_validation}

Before statistical analysis, we ran an automated validation pass over the PPO run corpus covering file integrity, numerical sanity, seed independence, reset-free semantics, family labels, recorded configuration labels, and cross-checks against main-text summary statistics. The pass checked recorded labels and values, not the settings that were executed, the code version, the hardware, or whether each threshold can act on each body (Appendix~\ref{sec:fallrules}).

\section{Detailed Reproducibility Analysis}
\label{sec:repro_appendix}

\paragraph{Body-alias lookup.}
Table~\ref{tab:body_aliases} maps the body aliases of Figures~\ref{fig:gait_chronophoto_hero},~\ref{fig:gait_temporal} and~\ref{fig:gait_chronophoto_full} to their full UNIMAL identifiers.

\begin{table}[h]
\centering
\caption{Aliases used in figure captions, and the full UNIMAL morphology identifiers.}
\label{tab:body_aliases}
\small
\begin{tabular}{@{}llp{0.55\linewidth}@{}}
\toprule
Alias & Family & UNIMAL morphology identifier \\
\midrule
Floor A & Floor & \texttt{floor-5506-10-6-01-15-48-35} \\
MVT A   & MVT   & \texttt{mvt-5506-13-10-17-12-25-45} \\
Floor B & Floor & \texttt{floor-1409-0-7-01-15-15-26} \\
\bottomrule
\end{tabular}
\end{table}

\subsection{Released Artifacts}
\label{sec:records}

The release is one code repository (\codeurl{}), with code under the MIT license and run records under CC BY 4.0 (Table~\ref{tab:artifacts}). One seeded CPU script recomputes 2,401 values from the run summaries without GPU training, among them nearly every main-text statistic of the UNIMAL panels, and its output lists the values it does not recompute. The code retrains the controllers of the 100 evolved bodies from scratch.

\begin{table}[htbp]
\centering
\caption{Contents of the release.}
\label{tab:artifacts}
\footnotesize
\setlength{\tabcolsep}{4pt}
\begin{tabular}{@{}p{0.24\linewidth}p{0.72\linewidth}@{}}
\toprule
Item & Status \\
\midrule
\appcell{Run summaries} & \appcell{Every reported UNIMAL panel except the advantage actor-critic (A2C) panel, and a panel on four existing legged robots, 3,306 runs in all} \\
\appcell{Additional panels} & \appcell{Per-run summaries of the 5,860 Humanoid, Ant and Evolution Gym runs with each panel's specification and frozen report. Their raw run records and training code are not released} \\
\appcell{Probe records} & \appcell{Zero-action rollouts and settle heights of all 100 bodies and the random-action searches of Appendix~\ref{sec:fallrules}, with the scripts that made all but one of them} \\
\appcell{Planned-test records} & \appcell{Specification, report and analysis output of each planned test except the A2C panel} \\
\appcell{Code} & \appcell{Training and launch code (the corpus-era trainer and the later revision) and CPU scripts that recompute the statistics and figures from the run summaries and probe records} \\
\appcell{Bodies} & \appcell{Morphology features and a script that downloads the UNIMAL body files from the upstream release and checks their hashes. The files are not redistributed, because the upstream repositories declare no license} \\
\appcell{Rollouts and logs} & \appcell{The traces behind Figures~\ref{fig:gait_chronophoto_hero}, \ref{fig:gait_temporal} and~\ref{fig:gait_chronophoto_full}, and training logs where kept (none for the corpus and SAC runs)} \\
\appcell{Not available} & \appcell{Policy checkpoints and the per-run code versions of the corpus and SAC runs, which were not saved, and the analyzed A2C runs (Appendix~\ref{sec:sac_appendix})} \\
\bottomrule
\end{tabular}
\end{table}

\section{Protocol Glossary and Per-Body Cutoff Checks}
\label{sec:protocol_glossary_appendix}
\label{sec:fallrules_rationale}

The three protocols in the main text are specified to differ only in the height-based early-termination threshold $\tau$ applied during training; all other episode mechanics are specified identically (Table~\ref{tab:protocol_glossary}).

\begin{table}[h]
\centering
\caption{Scope of Standard, Strict, and Reset-free as used in this paper. The column ``Early-termination on falls?'' refers to the height-based predicate during training; non-threshold episode mechanics are specified identically across protocols.}
\label{tab:protocol_glossary}
\small
\begin{tabular}{@{}p{0.17\linewidth}cp{0.20\linewidth}p{0.43\linewidth}@{}}
\toprule
Name in paper & $\tau$ & Early-termination on falls? & Literature anchor \\
\midrule
Standard   & $0.20$ & Yes, at 20\% of reset height & The default Gymnasium MuJoCo~\citep{towers2023gymnasium} height range roughly maps to an equivalent fraction of $\approx 0.27$ on an Ant with $z_0 \approx 0.75$~m (0.56, 0.64, and 0.71 of the initial torso height on Hopper, Walker2d, and Humanoid), all above 0.20. Our value is the fall rule of the locomotion environment our training code was adapted from. \\
Strict     & $0.50$ & Yes, at 50\% of reset height & DERL default~\citep{gupta2021embodied} sets a per-body ratio of 0.5, but applies it to its spawn height minus a 0.2\,m clearance and bootstraps at its time limit, so our cutoff is 0.1\,m higher; MetaMorph~\citep{gupta2022metamorph} uses the same configurable UNIMAL substrate. \\
Reset-free & $0$    & No early-termination on falls           & Fall-based resets are disabled during training, matching the endpoint of our height-threshold sweep. \\
\bottomrule
\end{tabular}
\end{table}

Throughout the paper, \emph{reset-free} denotes this height-threshold endpoint: the fall predicate is disabled, while the rest of the training setup is specified identically across protocols. Main-text body-level claims are scoped to this height-based termination mechanism. The geometry of Appendix~\ref{sec:fallrules} also appears in standard setups: DERL's own rule puts the cutoff of our 25 shortest bodies at half the torso radius, and Gymnasium Ant's lower bound of 0.2\,m lies below its own 0.25\,m torso radius, so the per-body check of Section~\ref{sec:cutoff_check} is worth running on standard setups too.

\subsection{Per-Body Cutoff Checks}
\label{sec:fallrules}

\paragraph{The Standard cutoff lies below the torso radius for 28 bodies.}
Every torso in the corpus is a sphere of radius 0.1\,m. For the 28 bodies with $z_0 \le 0.45$\,m (8 Floor, 13 VT, 7 MVT), the Standard cutoff $0.20\,z_0$ is 0.06 or 0.09\,m and lies below the torso radius. In simulation probes with random actions, flipped landings and drops, the torso of the 25 bodies with $z_0 = 0.30$\,m never reached the cutoff, and the torso of the three MVT bodies with $z_0 = 0.45$\,m went at most 5\,mm below it, so Standard training may have ended some of their early episodes, which the corpus did not record. All 15 policies of each of the 28 bodies produce the same record under the common evaluator as under the neutral one, so Standard training is in effect Reset-free training for these bodies, whose Reset-free to Standard ratio averages 0.98. The same geometry limits the threshold sweep: the $\tau = 0.10$ cutoff lies below the torso radius for 94 bodies, and the $\tau = 0.35$ cutoff is reached whenever the torso of one of the 25 shortest bodies rests on the floor.

\paragraph{The Strict cutoff can be crossed during the drop from the reset pose.}
Because $z_0$ is a spawn height, the Strict cutoff $0.5\,z_0$ can lie above the height at which a body comes to rest, and with zero action 45 of the 100 bodies are below it after 1\,s. A Strict episode on such a body continues only if the controller catches the drop, and most controllers learn to. We call a Strict-trained run cut off if its episodes under the Strict evaluator average at most 20 steps. Cut-off runs are 17 of 123 Floor, 31 of 111 VT and 1 of 66 MVT Strict runs, and they account for most of the Strict loss of Floor and VT bodies (about 15 of 22 and 28 of 34 points, post hoc). For ten VT bodies, every PPO controller we trained averages at most 26 steps under the Strict evaluator, while SAC keeps one of them up. An eleventh body, a Floor body with $z_0 = 0.75$\,m, has cut-off Strict-trained runs in every seed, which this geometry does not explain. We call these 11 bodies, whose Strict-trained runs are cut off on average, the cut-off bodies, and the 23 bodies with at least one cut-off seed the cut-off-seed bodies. Both sets are defined by outcomes, so results without them are post hoc sensitivity checks. The numbers record of the release (\texttt{records/camera\_ready\_numbers.json}) lists the affected body sets by identifier, and its probe records hold the probe details (Table~\ref{tab:artifacts}).

\section{Evaluation-Protocol Robustness}
\label{sec:regret_robustness_appendix}

\paragraph{Selection-retention robustness.}
The main-text selection analysis (Section~\ref{sec:regret}) scores every policy at $\tau_{\text{eval}} = 0.20$, and Table~\ref{tab:regret_robustness} reports the same retention at $\tau_{\text{eval}} \in \{0.00, 0.20, 0.50\}$. The common and neutral rows agree closely, partly because the two evaluators coincide for the 28 bodies whose Standard cutoff lies below the torso radius: top-5 Standard-selected retention against the Strict oracle is $49.0\%$ and $49.3\%$, with overlapping confidence intervals. The $\tau_{\text{eval}} = 0.50$ rows are diagnostic only, because Standard rewards approach zero for many bodies under that evaluator.

\begin{table}[h]
\centering
\caption{Selection retention (target-training oracle retained by a Standard-training selection) across evaluation thresholds. Morphology bootstrap with $B = 10{,}000$ resamples. The $\tau_{\text{eval}} = 0.20$ rows match Section~\ref{sec:regret}; the $\tau_{\text{eval}} = 0$ rows are the threshold-free robustness check. The $\tau_{\text{eval}} = 0.50$ rows are diagnostic because Standard denominators approach zero.}
\label{tab:regret_robustness}
\small
\begin{tabular}{lcrr}
\toprule
$\tau_{\text{eval}}$ & $k$ & vs.\ Strict oracle $\uparrow$ & vs.\ RF oracle $\uparrow$ \\
\midrule
0.20 & 5  & 49.0\% [22.8, 93.5] & 95.1\% [90.9, 100.0] \\
0.20 & 10 & 74.9\% [50.1, 93.3] & 96.4\% [92.3, 100.0] \\
0.20 & 20 & 85.1\% [66.9, 93.0] & 94.4\% [83.9, 100.0] \\
\midrule
0.00 & 5  & 49.3\% [23.0, 93.5] & 96.3\% [92.5, 100.0] \\
0.00 & 10 & 75.1\% [50.0, 92.6] & 96.9\% [94.3, 100.0] \\
0.00 & 20 & 83.7\% [65.6, 92.7] & 94.8\% [85.8, 99.9] \\
\midrule
0.50 & 5  & 89.6\% [78.3, 99.0] & 100.0\% [56.2, 100.0] \\
0.50 & 10 & 93.9\% [85.8, 98.6] & 75.8\% [51.1, 100.0] \\
0.50 & 20 & 90.7\% [84.1, 96.1] & 73.3\% [62.7, 89.8] \\
\bottomrule
\end{tabular}
\end{table}

\begin{figure}[h]
    \centering
    \includegraphics[width=\textwidth]{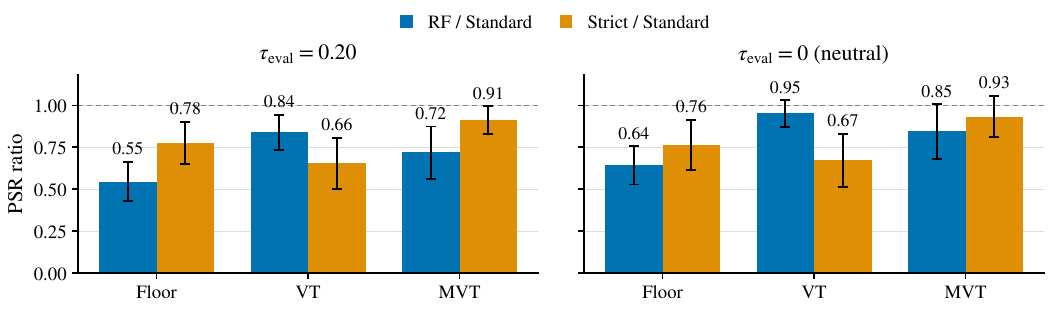}
    \caption{Family-level RF/Std and Str/Std PSR ratios at $\tau_{\text{eval}} = 0.20$ and neutral $\tau_{\text{eval}} = 0$. Bars show family means over per-body ratios, with 95\% morphology-bootstrap CIs. The Floor and VT difference in Str/Std comes mostly from runs cut off during the drop from the reset pose (Appendix~\ref{sec:fallrules}). \textbf{Takeaway:} the Reset-free family ordering persists under neutral evaluation, so the evaluation cutoff does not produce it.}
    \label{fig:robustness_eval}
\end{figure}

\FloatBarrier
\section{Protocol-Robust Selection: Full Retention Matrix}
\label{sec:robust_selection_appendix}

Table~\ref{tab:robust_selection_full} gives the full criterion-by-oracle-by-$k$ matrix with bootstrap 95\% CIs ($B = 10{,}000$ body resamples). Min-over-protocols is the only criterion with point estimates above $90\%$ against every oracle at every $k$ ($92.3$/$92.8$/$91.4\%$), and mean-over-protocols is near $90\%$ throughout ($92.3$/$93.9$/$89.9\%$). At $k=5$ the two select the same top-5 set, so their point retentions coincide. All values are in sample, because the same seed-averaged scores choose the bodies and define the oracles. The Standard-only $49.0\%$ against the Strict oracle at $k = 5$ comes mostly from the bodies whose cutoffs do not act as intended: it is $75.7\%$ on the 72 bodies whose Standard cutoff lies above the torso radius.

\begin{table}[h]
\centering
\caption{In-sample retention of each single-protocol oracle's top-$k$ PSR under five selection criteria, at common $\tau_{\text{eval}} = 0.20$. Point estimates with bootstrap 95\% CIs. Oracle diagonals are $100\%$ by construction. $^\dagger$ marks our conservative selector.}
\label{tab:robust_selection_full}
\small
\begin{tabular}{llccc}
\toprule
Criterion & $k$ & vs.\ Standard oracle $\uparrow$ & vs.\ Strict oracle $\uparrow$ & vs.\ Reset-free oracle $\uparrow$ \\
\midrule
Standard-only        & 5  &  n/a                     & 49.0\% [22.8, 93.5] & 95.1\% [90.9, 100.0] \\
                     & 10 &  n/a                     & 74.9\% [50.1, 93.3] & 96.4\% [92.3, 100.0] \\
                     & 20 &  n/a                     & 85.1\% [66.9, 93.0] & 94.4\% [83.9, 100.0] \\
\midrule
Strict-only          & 5  & 91.1\% [81.6, 97.2]      &  n/a                & 93.3\% [74.6, 99.2] \\
                     & 10 & 90.4\% [81.3, 97.0]      &  n/a                & 87.9\% [61.7, 97.7] \\
                     & 20 & 90.1\% [84.6, 95.5]      &  n/a                & 77.4\% [60.2, 90.6] \\
\midrule
Reset-free-only      & 5  & 95.6\% [92.5, 100.0]     & 67.7\% [36.5, 97.9] &  n/a                \\
                     & 10 & 98.9\% [95.6, 100.0]     & 81.9\% [60.1, 94.2] &  n/a                \\
                     & 20 & 98.9\% [96.7, 100.0]     & 78.6\% [60.9, 90.5] &  n/a                \\
\midrule
Mean over protocols  & 5  & 92.3\% [86.9, 99.0]      & 99.1\% [88.5, 100.0] & 95.7\% [88.1, 100.0] \\
                     & 10 & 96.7\% [91.9, 99.9]      & 93.9\% [80.9, 100.0] & 95.8\% [91.0, 99.5] \\
                     & 20 & 97.9\% [94.0, 99.7]      & 89.9\% [82.7, 96.9] & 98.4\% [93.5, 99.8] \\
\midrule
Min over protocols$^\dagger$ & 5  & 92.3\% [84.3, 97.9] & 99.1\% [98.2, 100.0] & 95.7\% [85.2, 99.6] \\
                     & 10 & 93.5\% [85.2, 97.9] & 99.4\% [96.6, 100.0] & 92.8\% [85.2, 98.3] \\
                     & 20 & 91.4\% [85.4, 96.3] & 96.7\% [92.1, 99.7] & 92.1\% [84.7, 96.9] \\
\bottomrule
\end{tabular}
\end{table}

\textbf{Held-out seed.} To gauge how optimistic the in-sample values are, we also choose bodies on the mean of two seeds and score them against the third seed's own top $k$, averaging over the three held-out seeds (post hoc). Min-over-protocols keeps 82.8\%, 84.3\% and 85.2\% at $k = 5$, 10 and 20, the highest worst-case retention of the five criteria at every $k$ (the next best, the mean over protocols, keeps 82.2\%, 78.9\% and 81.6\%). Of the paired body-bootstrap intervals for its differences from Standard-only, Strict-only and mean-over-protocols, only those against Standard-only at $k = 5$ and 20 and against Strict-only at $k = 20$ exclude zero.

\textbf{Held-out protocol.} A planned test held out each protocol in turn. It required that, at $k = 5$, the best criterion built from the other two protocols beat the better of those two single protocols on the held-out one by at least 0.05 PSR in two of three splits, with an interval above zero, and that some combination keep at least 85\% of the held-out top-10 reward in every split. Neither part passed: the gains are $+0.02$, $-0.13$ and $+0.23$ PSR (mean $+0.04$, $[-0.42, 0.23]$), and with Strict held out the best combination keeps 81.9\% of the held-out top-10 reward. The test therefore does not support the selector as a predictor for an untested protocol.

\textbf{Family composition under each criterion.} The top-10 picks under each selection criterion have the following family split (Floor / VT / MVT): Standard-only $2/2/6$; Strict-only $2/1/7$; Reset-free-only $2/2/6$; mean $2/0/8$; min $2/0/8$. The two protocol-robust criteria shift the MVT count from $6$--$7$ to $8$ without requiring family labels as input. Five of the eight MVT bodies in the min-over-protocols top 10 are among the 28 bodies whose Standard cutoff lies below the torso radius, and on the other 72 bodies the mean and min criteria each pick 4 MVT bodies.

\section{Continuous-Threshold Sweep}
\label{sec:threshold_landscape_appendix}

\begin{figure}[h]
    \centering
    \includegraphics[width=0.85\textwidth]{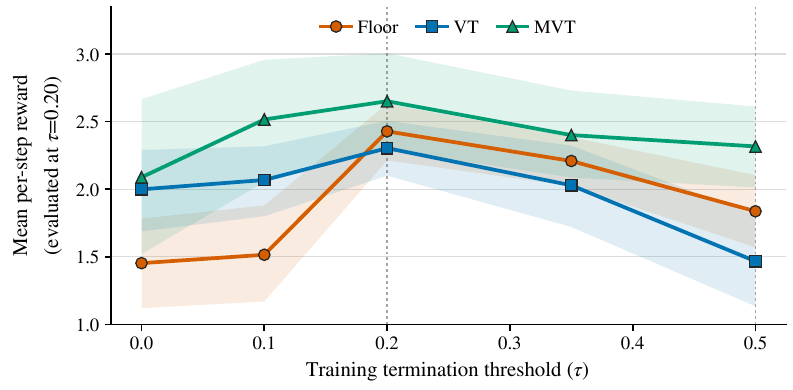}
    \caption{Mean per-step reward by training threshold, per family. Shaded ribbons are 95\% morphology-bootstrap CIs ($B=5{,}000$) computed on seed-averaged morphologies, with each policy evaluated at $\tau_{\text{eval}} = 0.20$. VT's drop at strict termination comes mostly from the ten VT bodies whose Strict runs are cut off during the drop from the reset pose. The $\tau = 0.10$ and $0.35$ points show the shape of reward across $\tau$ only, because those cutoffs do not act on every body as their values suggest. \textbf{Takeaway:} Floor bodies lose the most without fall termination.}
    \label{fig:threshold_curves}
\end{figure}

Figure~\ref{fig:threshold_curves} shows the mean per-step reward (seed-averaged, evaluated at $\tau_{\text{eval}} = 0.20$) as a function of the training termination threshold $\tau$ at five measured points ($\tau \in \{0.00, 0.10, 0.20, 0.35, 0.50\}$), per evolutionary family. The discrete main-text comparison at $\tau \in \{0, 0.20, 0.50\}$ samples the same threshold response: Floor peaks sharply at $\tau = 0.20$ (2.43) and drops steeply toward reset-free ($\tau = 0$, $1.45$), VT drops steeply toward strict ($\tau = 0.50$, $1.47$, or $2.01$ without the ten VT bodies whose Strict runs are cut off during the drop from the reset pose), and MVT's curve is the flattest (range $[2.09, 2.65]$ across all five thresholds). The intermediate thresholds do not act as their values suggest, because the $\tau = 0.10$ cutoff lies below the torso radius of 94 bodies and the $\tau = 0.35$ cutoff is reached whenever the torso of one of the 25 shortest bodies rests on the floor (Appendix~\ref{sec:fallrules}). We therefore read the sweep for the shape of reward across $\tau$ only, not as showing the same family response at every threshold.

\section{Within-Family Ratio Distributions}
\label{sec:ratio_distributions_appendix}

\begin{figure}[h]
    \centering
    \includegraphics[width=\textwidth]{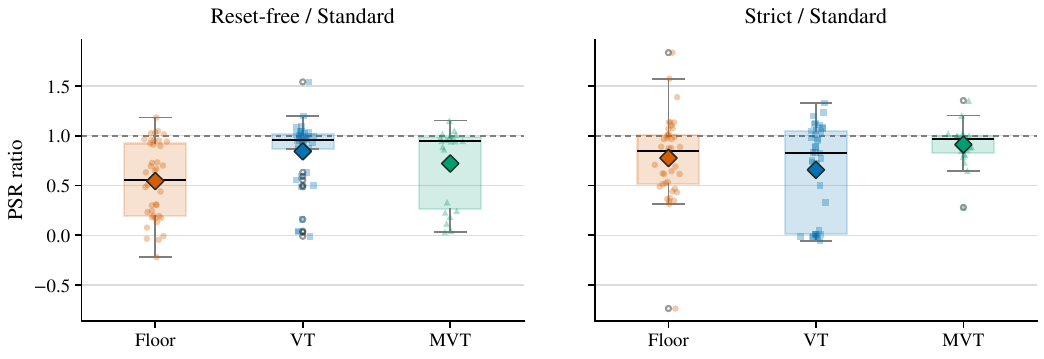}
    \caption{Per-morphology RF/Std (left) and Str/Std (right) ratios by family, using seed-averaged PSR at $\tau_{\text{eval}} = 0.20$. Boxes show medians and IQRs, and diamonds family means. The low VT Str/Std values come mostly from runs cut off during the drop from the reset pose (Appendix~\ref{sec:fallrules}). \textbf{Takeaway:} the family difference in RF/Std is a shift of the whole distribution, not an outlier effect.}
    \label{fig:ratio_distributions}
\end{figure}

The family-level means reported in Table~\ref{tab:family} average over heterogeneous individual morphologies. Figure~\ref{fig:ratio_distributions} shows the full within-family distribution of RF/Std and Str/Std ratios. Every Standard denominator is safely positive (minimum seed-averaged Standard PSR $=0.789$), so the ratios are not artifacts of near-zero baselines.

\section{Per-Seed Replication and Alternative Blocked Models}
\label{sec:per_seed_appendix}

\subsection{Robustness of the Protocol-by-Family Interaction}
\label{sec:per_seed_robustness}

Table~\ref{tab:interaction_robustness} adds five robustness checks to the per-seed ANOVA of the protocol-by-family interaction. The ANOVAs on per-seed rows treat a body's three seeds as independent, so they are supplementary, and the tests that treat the body as the unit are primary. All five confirm the interaction direction: four return $p < 0.01$, and the mixed-effects model returns a sign-consistent VT-by-reset-free coefficient with a 95\% CI excluding zero. Without the 11 bodies whose Strict runs are cut off on average, the permutation $p$ is 0.14 (Appendix~\ref{sec:ranking_appendix}) and the clustered Wald $p$ is 0.007.

\begin{table}[h]
\centering
\caption{Protocol-by-family interaction: supplementary per-seed ANOVA plus five robustness checks (Section~\ref{sec:results}). The mixed-effects row reports the VT-by-reset-free interaction coefficient with 95\% CI in place of a $p$-value.}
\label{tab:interaction_robustness}
\small
\begin{tabular}{@{}p{0.34\linewidth}p{0.36\linewidth}p{0.24\linewidth}@{}}
\toprule
Test & Statistic & $p$-value \\
\midrule
Per-seed two-way ANOVA              & $F(4, 794) = 12.54$          & $6.8 \times 10^{-10}$ \\
Morphology-level permutation        & $F_{\text{obs}} = 12.54$ vs.\ null 95th $= 7.39$ & $5.2 \times 10^{-3}$ ($B{=}5{,}000$); $4.5 \times 10^{-3}$ ($B{=}10{,}000$) \\
Seed-averaged two-way ANOVA         & $F(4, 194) = 4.79$           & $1.1 \times 10^{-3}$ \\
Rank-transformed per-seed ANOVA     & $F(4, 794) = 11.43$          & $5.1 \times 10^{-9}$ \\
Cluster-robust Wald + GEE backup    & $F(4, 99) = 4.79$; $\chi^2_4 = 21.91$ & $1.4 \times 10^{-3}$; $2.1 \times 10^{-4}$ \\
Mixed-effects (random morph intercept) & VT$\times$RF $= +0.67$, CI $[0.39, 0.95]$ & n/a \\
\bottomrule
\end{tabular}
\end{table}

\subsection{Balanced-Family Robustness}
\label{sec:per_seed_balanced}

Family sizes are imbalanced (Floor $n = 41$, VT $n = 37$, MVT $n = 22$), so we repeated the PSR interaction analysis on 1,000 balanced subsamples with 22 morphologies per family (deterministic seed). The directional family ordering is stable under this reanalysis: Floor RF/Std remains below VT RF/Std in all 1,000 balanced draws.

\subsection{Leave-One-Family-Out Robustness}
\label{sec:per_seed_lofo}

To identify which families drive the interaction we refit the per-seed blocked ANOVA on every 2-family subset. The two anchored contrasts are stable in this per-seed test, which treats seeds as independent: dropping MVT preserves the Floor-vs-VT contrast ($F(2,620) = 21.01$, $p = 1.5 \times 10^{-9}$, body-level permutation $p < 0.01$), and dropping Floor preserves the VT/MVT strict-termination contrast ($F(2,468) = 11.09$, $p = 2.0 \times 10^{-5}$, body-level permutation $p = 0.043$). The Floor-vs-VT contrast on per-body RF/Std (Floor $0.55$ vs.\ VT $0.84$) has $\Delta = -0.298$, 95\% CI $[-0.454, -0.137]$ ($B = 10{,}000$, bootstrap within families), and the VT-vs-MVT contrast on Str/Std (VT $0.66$ vs.\ MVT $0.91$) has $\Delta = -0.254$, $[-0.430, -0.082]$. Of these two contrasts, only the Floor-vs-VT contrast in RF/Std is a family result, because the VT-vs-MVT Strict contrast comes from the ten VT bodies whose Strict runs are cut off during the drop from the reset pose: without them it is $-0.01$, $[-0.15, 0.12]$ (Appendix~\ref{sec:fallrules}).

\subsection{Per-Seed RF/Std ANOVA}
\label{sec:per_seed_rf_std}

\begin{figure}[h]
    \centering
    \includegraphics[width=\textwidth]{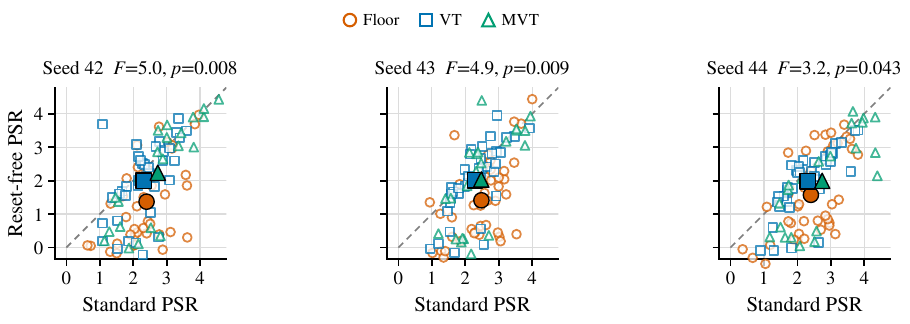}
    \caption{Per-seed scatter: Standard vs.\ Reset-free per-step reward for each random seed. Larger filled markers show family means. \textbf{Takeaway:} the family separation is consistent across seeds.}
    \label{fig:per_seed}
\end{figure}

The family interaction replicates across all three random seeds (Figure~\ref{fig:per_seed} and Table~\ref{tab:per_seed}).

\begin{table}[h]
\centering
\caption{Per-seed replication of the family interaction. The last row averages rewards over seeds before forming ratios, as in Table~\ref{tab:family}.}
\label{tab:per_seed}
\small
\begin{tabular}{lccccc}
\toprule
Seed & Floor RF/Std $\uparrow$ & VT RF/Std $\uparrow$ & MVT RF/Std $\uparrow$ & $F$ & $p$ \\
\midrule
42 & 0.528 & 0.875 & 0.747 & 5.02 & 0.008 \\
43 & 0.545 & 0.846 & 0.773 & 4.90 & 0.009 \\
44 & 0.594 & 0.833 & 0.679 & 3.24 & 0.043 \\
\midrule
Seed-averaged & 0.545 & 0.843 & 0.722 & 6.46 & 0.002 \\
\bottomrule
\end{tabular}
\end{table}

The RF/Std ordering (VT $>$ MVT $>$ Floor) holds in all three seeds; per-seed ANOVA statistics range from $F=3.24$ to $5.02$ ($p \in [0.008,0.043]$). The seed-averaged ANOVA on RF/Std ratios ($F = 6.46$, $p = 0.002$) and the bootstrap confidence interval for the VT-Floor difference ($[0.14, 0.45]$) confirm the robustness of the effect.

\subsection{Alternative Blocked Models}
\label{sec:per_seed_alt_models}

The main-text per-seed interaction test, which is supplementary because it treats a body's three seeds as independent, uses a per-seed blocked ANOVA ($n = 900$) with morphology fixed effects, protocol, and the protocol-by-family interaction; morphology is nested in family, so the family main effect is absorbed and not interpreted. The interaction gives $F(4, 794) = 12.54$, $p = 6.8 \times 10^{-10}$, partial $\eta^2 = 0.059$. Two alternative specifications preserve the interaction:
\begin{itemize}
    \item \textbf{Seed-averaged blocking} (one observation per cell after averaging across the three seeds per (morphology, protocol); $n = 300$): interaction $F(4, 194) = 4.79$, $p = 0.00106$, partial $\eta^2 = 0.090$; protocol main $F(2, 194) = 25.31$, $p = 1.7 \times 10^{-10}$. Each body belongs to one family, so once each body has its own effect a family main effect is not identifiable, and we do not report one.
    \item \textbf{Mixed-effects random-intercept model} with protocol, family, their interaction, and a random intercept per morphology on per-seed data recovers the main reset-free contrast: reset-free main $= -0.975$ (95\% CI $[-1.17, -0.78]$), which is the effect in the Floor reference family (the average over all bodies is $-0.64$), and VT-by-reset-free interaction $= +0.670$ (95\% CI $[0.39, 0.95]$). We draw no strict-side family conclusion from these models, because the strict-side family contrast comes from runs cut off during the drop from the reset pose (Appendix~\ref{sec:fallrules}).
\end{itemize}

\subsection{Morphology-Level Permutation Test}
\label{sec:per_seed_permutation}

The parametric $F$-distribution on the blocked model could in principle be unreliable because morphology is nested in family. We therefore add a non-parametric permutation complement: shuffle family labels across the 100 morphologies while preserving family sizes (41, 37, 22), refit the same blocked two-way ANOVA, and record the Type II interaction $F$. Under the null that protocol response is family-independent, the observed $F = 12.54$ should fall within the permutation $F$ distribution; instead it exceeds the 95th percentile ($7.4$) of the null and yields $p_{\text{perm}} = 0.005$ at $B = 5{,}000$ (null mean $= 2.75$, null max $= 19.02$), and the released recomputation gives $p_{\text{perm}} = 0.004$ at $B = 10{,}000$. Without the 11 bodies whose Strict runs are cut off on average, the permutation $p$ is 0.14 (Appendix~\ref{sec:ranking_appendix}). A simpler one-way ANOVA on per-morphology RF/Std ratios (no morph block, so no nesting concern) gives $F = 6.46$ with $p_{\text{perm}} = 0.003$ at $B = 10{,}000$ (null 95th percentile $= 3.06$). Both permutation tests confirm the interaction is not an artifact of the parametric $F$ distribution or the nested design.

\subsection{Metric Sensitivity: Per-Step Reward vs.\ Total-Reward Per Horizon}
\label{sec:per_seed_metric}

Per-step reward (PSR) estimates a reward rate; total reward over the fixed horizon $T_{\max}{=}1000$ ($\text{fitness} / T_{\max}$) treats post-termination reward as zero and is the natural conservative complement when episodes end early. PSR can over-credit a fast-then-fall policy by averaging only over surviving steps; $\text{fitness}/T_{\max}$ cannot. Recomputing every Table~\ref{tab:family} ratio under $\text{fitness}/T_{\max}$ gives RF/Std Floor $0.473$, VT $0.847$, MVT $0.678$ (vs.\ PSR ratios $0.545$, $0.843$, $0.722$), and the one-way family ANOVA on fitness RF/Std yields $F(2, 97) = 8.37$, $p = 4.4 \times 10^{-4}$, larger than the PSR-based $F = 6.46$. The per-body PSR-vs-fitness rank correlation exceeds $0.95$ at every protocol. The family difference in RF/Std does not depend on the PSR normalization.

\subsection{Compute-Budget Audit: Where the 5M Steps Go}
\label{sec:compute_budget_appendix}
\label{sec:per_seed_compute}

A natural concern is that under reset-free training (no fall termination), Floor bodies might consume the 5M-step budget in unrecoverable post-fall states, so the Floor RF underperformance would reflect an effective compute deficit rather than a body-by-protocol learning effect. The trained policies argue against this. Under neutral evaluation ($\tau_{\text{eval}} = 0$, no z-fall termination), every (family, protocol) cell averages at least 999 of 1000 steps (Floor RF $= 1000.0$; Floor Strict $= 999.0$; all other cells $= 1000.0$), because without a fall clause only the upper height bound can end an episode early, and it did so for one controller only. Eval-time fall behavior at the paper's headline threshold ($\tau_{\text{eval}} = 0.20$) shows Floor RF drops below $0.20\,z_0$ in $61\%$ of evaluation episodes (mean episode length $512$ vs.\ $828$ for Floor Standard), but neutral-eval distance for Floor RF is $31.2\,\text{m}$, comparable to Floor Strict ($31.2$) and below Floor Standard ($44.5$). Floor RF is therefore not simply a collapsed policy: under neutral evaluation it keeps moving, but its torso often drops below the standard height threshold.

\section{Morphological Features Track Sensitivity}
\label{sec:features_appendix}

The families differ in evolutionary origin, so we also ask whether structural features alone track protocol sensitivity in this corpus. We correlate simple graph-derived morphology descriptors (tree depth, joint count, rigid-body count, first-level limb count, and reset height) with RF/Std and Str/Std PSR ratios (Table~\ref{tab:features}).

\begin{table}[h]
\centering
\caption{Spearman correlations between representative morphological features and PSR sensitivity ratios. All correlations are computed on per-morphology RF/Std and Str/Std ratios.}
\label{tab:features}
\small
\begin{tabular}{lcccc}
\toprule
 & \multicolumn{2}{c}{RF/Std} & \multicolumn{2}{c}{Str/Std} \\
Feature & $\rho$ & $p$ & $\rho$ & $p$ \\
\midrule
Initial height      & $-$0.533 & $<$0.001 & 0.382 & $<$0.001 \\
Num.\ first-level limbs & 0.496  & $<$0.001 & $-$0.370 & $<$0.001 \\
Tree depth          & $-$0.483 & $<$0.001 & 0.403 & $<$0.001 \\
Num.\ joints        & $-$0.117 & 0.246  & 0.480 & $<$0.001 \\
Num.\ bodies        & $-$0.004 & 0.965  & 0.375 & $<$0.001 \\
\bottomrule
\end{tabular}
\end{table}

Initial height and number of first-level limbs are the clearest opposite-sign predictors. Taller bodies are less robust to reset-free training ($\rho = -0.533$) but more robust to strict termination ($\rho = 0.382$). More first-level limbs show the reverse pattern ($\rho = 0.496$ for RF/Std, $\rho = -0.370$ for Str/Std). Both survive Bonferroni correction across the feature-ratio correlations in Table~\ref{tab:features} on all 100 bodies. The Str/Std correlations depend on the 28 shortest bodies, and none survives the correction on the 72 bodies whose Standard cutoff lies above the torso radius (Appendix~\ref{sec:fallrules}), where height has $\rho = 0.03$ ($p = 0.81$). These feature-wise correlations are partially confounded with family labels (because families differ systematically on these features); the linear model and leave-one-family-out cross-validation below probe whether a family-controlled structural predictor remains informative.

\paragraph{Linear feature model for protocol sensitivity.}
We fit an ordinary least squares regression of the paired delta $\Delta_{\text{str-rf}}(m) = \text{PSR}_{\text{strict}}(m) - \text{PSR}_{\text{RF}}(m)$ on the same five structural features across all 100 corpus morphologies. In-sample $R^2 = 0.46$; leave-one-out cross-validated Spearman $\rho = 0.653$ ($p = 1.9 \times 10^{-13}$). Standardized coefficients (per standard-deviation feature increment) are: initial height $+0.460$, tree depth $+0.352$, number of joints $+0.245$, number of first-level limbs $-0.167$, number of bodies $-0.137$. Four of the five coefficients agree in sign with the Spearman correlations (the rigid-body coefficient does not, because the number of rigid bodies is collinear with the number of joints, $r = 0.84$), and they show that taller, deeper bodies tend to be worse under reset-free training (positive $\Delta_{\text{str-rf}}$), while wider, more-limbed bodies tend the other way.

\paragraph{Are structural features doing work beyond baseline skill?}
A simpler explanation is that the feature model recapitulates overall learnability: a high-skill body wins under any protocol, so $\Delta_{\text{str-rf}}$ is just a function of $\text{PSR}_{\text{Std}}$. To check, we fit three nested models on the 100-body corpus (Table~\ref{tab:structural_models}): M1 uses only baseline Standard skill, M2 only structural features, and M3 both. If the effect were purely a skill artifact, M1 would dominate and M3 would equal M1. Instead, M2 alone explains four times as much variance as M1 ($R^2 = 0.46$ vs.\ $0.11$) and achieves Spearman $\rho = 0.71$ (vs.\ $0.28$); M3 adds only small lift over M2 ($R^2 = 0.55$, $\rho = 0.74$). In this corpus, structural features carry information about protocol sensitivity that baseline skill does not, and the two signals are largely complementary.

A smaller two-feature control using only initial height and first-level limb count explains $R^2 = 0.42$ of $\Delta_{\text{str-rf}}$. Its family residual means remain opposite-signed (Floor $+0.177$, VT $-0.342$, MVT $+0.244$; one-way ANOVA $F=3.32$, $p=0.04$, against $F = 5.12$, $p = 0.008$ for the families before the control), so these two structural features reduce but do not erase the family pattern.

\begin{table}[h]
\centering
\caption{In-sample fit for three nested predictors of $\Delta_{\text{str-rf}} = \text{PSR}_{\text{strict}} - \text{PSR}_{\text{RF}}$ on the 100 UNIMAL corpus bodies. M2 features: initial height, num.\ first-level limbs, tree depth, num.\ joints, num.\ bodies. Spearman 95\% CIs are bootstrap ($B{=}10{,}000$) over morphologies.}
\label{tab:structural_models}
\small
\begin{tabular}{llccc}
\toprule
Model & Predictors & $R^2 \uparrow$ & AIC $\downarrow$ & Spearman $\rho \uparrow$ [95\% CI] \\
\midrule
M1 & $\text{PSR}_{\text{Std}}$                  & 0.11 & 54.3 & 0.28 [$0.08$, $0.45$] \\
M2 & structural features (5)                    & 0.46 & 11.1 & 0.71 [$0.59$, $0.79$] \\
M3 & $\text{PSR}_{\text{Std}}$ + structural (6) & 0.55 & $-3.4$ & 0.74 [$0.63$, $0.82$] \\
\bottomrule
\end{tabular}
\end{table}

\paragraph{Leave-one-family-out cross-validation.}
To test whether the structural-feature signal transfers across DERL lineages within UNIMAL, we hold out one family, fit M1/M2/M3 on the other two, and evaluate the held-out family (Table~\ref{tab:lofo_structural}). M2 achieves Spearman point-estimate $\rho \geq 0.48$ on all three held-out families with 95\% CIs excluding zero. The transfer is strongest on the smaller MVT fold and remains positive on Floor and VT, but reset height alone ranks each held-out family at least as well ($\rho$ of 0.62, 0.49, and 0.79), so we treat the structural model as a description of this corpus rather than a predictor.

\begin{table}[h]
\centering
\caption{Leave-one-family-out cross-validation. For each held-out family, M1/M2/M3 are fit on the other two families' bodies, and Spearman $\rho$ between predicted and observed $\Delta_{\text{str-rf}}$ is reported with bootstrap 95\% CIs ($B{=}10{,}000$, resampling held-out bodies). Reset height alone ranks each held-out family at least as well as M2.}
\label{tab:lofo_structural}
\small
\begin{tabular}{lccccccc}
\toprule
Held-out family & $n_{\text{test}}$ & \multicolumn{2}{c}{M1} & \multicolumn{2}{c}{M2} & \multicolumn{2}{c}{M3} \\
\cmidrule(lr){3-4} \cmidrule(lr){5-6} \cmidrule(lr){7-8}
 & & $\rho \uparrow$ & 95\% CI & $\rho \uparrow$ & 95\% CI & $\rho \uparrow$ & 95\% CI \\
\midrule
Floor & 41 & 0.19 & [$-0.13$, $0.48$] & 0.51 & [$0.21$, $0.73$] & 0.53 & [$0.26$, $0.73$] \\
VT    & 37 & 0.36 & [$-0.00$, $0.64$] & 0.48 & [$0.17$, $0.69$] & 0.60 & [$0.32$, $0.78$] \\
MVT   & 22 & 0.69 & [$0.36$, $0.89$]   & 0.76 & [$0.42$, $0.93$] & 0.80 & [$0.49$, $0.94$] \\
\bottomrule
\end{tabular}
\end{table}

\section{Cross-Algorithm Validation with SAC}
\label{sec:sac_appendix}

As an optimizer-robustness check, we trained SAC~\citep{haarnoja2018soft} (the Brax implementation~\citep{freeman2021brax}) on a 24-morphology subset ($6$ Floor, $13$ VT, $5$ MVT) at $\tau \in \{0.00, 0.20, 0.50\}$ across three seeds for $5 \times 10^6$ environment steps each, giving a $3 \times 3$ cell per body and $n = 216$ total runs. The bodies were fixed before any SAC analysis to cover three outcomes under Strict training: 8 that fell often, 8 that stayed upright but barely moved, and 8 that were unaffected. The panel therefore covers the range of outcomes rather than estimating family averages. Its evaluation differs from PPO's in three details: SAC is scored with its deterministic action, its reset noise perturbs positions but not velocities, and its reset draws vary by seed.

The panel met its validation gate, a per-run protocol-by-family interaction with $p < 0.01$ and the Floor $>$ VT $>$ MVT ordering on the Strict to Reset-free ratio, which was set after a 96-run pilot of Reset-free and Strict training had shown that ordering.

\begin{table}[h]
\centering
\caption{SAC per-step reward by family (seed-averaged across three SAC seeds, eval at $\tau_{\text{eval}} = 0.20$) and the Str/RF validation ratio. Str/RF averages $\text{PSR}_{\text{strict}} / \text{PSR}_{\text{RF}}$ across bodies in each family. Values above 1 favor strict termination.}
\label{tab:sac_family}
\small
\begin{tabular}{lcrrrc}
\toprule
Family & $n$ & Standard $\uparrow$ & Strict $\uparrow$ & Reset-free $\uparrow$ & Str/RF \\
\midrule
Floor & 6  & 3.92 & 4.34 & 2.80 & 2.41 \\
VT    & 13 & 3.43 & 3.10 & 2.97 & 1.52 \\
MVT   & 5  & 4.07 & 3.92 & 4.10 & 0.96 \\
\bottomrule
\end{tabular}
\end{table}

The planned family-ordering check, Floor $>$ VT $>$ MVT on per-body Str/RF, holds under SAC ($2.41, 1.52, 0.96$) for the mean of per-body ratios on these 24 bodies (Table~\ref{tab:sac_family}), and SAC gives VT bodies a smaller Strict penalty than PPO (VT Strict minus Reset-free PSR $+0.12$ under SAC against $-0.51$ under PPO). SAC changes some absolute magnitudes, so we interpret the panel as a Str/RF ordering check rather than cross-algorithm replication of all PPO ratios. Cross-algorithm rank agreement is strong on the 23 bodies with denominator-valid PPO and SAC Str/RF ratios: Spearman $\rho = 0.75$ ($p = 3.1 \times 10^{-5}$). The omitted body has nonpositive PPO reset-free PSR, so its PPO Str/RF denominator fails the pre-specified denominator threshold (PSR $>0.05$); all SAC cells are valid. Leave-one-body-out sensitivity gives min $\rho = 0.720$, median $0.761$, max $0.833$.

A blocked ANOVA on per-seed SAC PSR ($n = 216$) with body fixed effects, protocol, and the protocol-by-family interaction gives $F(4, 186) = 6.71$ ($p = 4.6 \times 10^{-5}$), matching the main-text per-seed specification, which treats a body's seeds as independent. Tests that treat the body as the unit give $p = 0.016$ (clustered Wald), $0.070$ (seed-averaged), and $0.09$ (permutation of family labels). Body is nested in family, so the family main effect is absorbed rather than interpreted. We use this as cross-algorithm support for the Str/RF ordering; the PPO corpus remains the source for RF/Std magnitudes and percentage-loss claims.

\paragraph{Quality checks.}
216/216 cells are valid (no NaN). Median seed-level standard deviations are $\sigma_{\text{RF}} = 0.148$, $\sigma_{\text{Std}} = 0.181$, $\sigma_{\text{strict}} = 0.257$, all below the planned reliability threshold ($0.514$).

\paragraph{A2C.}
We also set up synchronous A2C on the same design (24 bodies, three rules and three seeds, 216 runs) under the same gate. The analyzed runs are no longer available, and the surviving record of the analysis reports an interaction $p = 0.073$, above the gate's 0.01.

\Needspace{0.62\textheight}%
\section{Morphology Gallery}
\label{sec:morph_gallery_appendix}

\begin{figure}[H]
    \centering
    \includegraphics[width=0.85\textwidth]{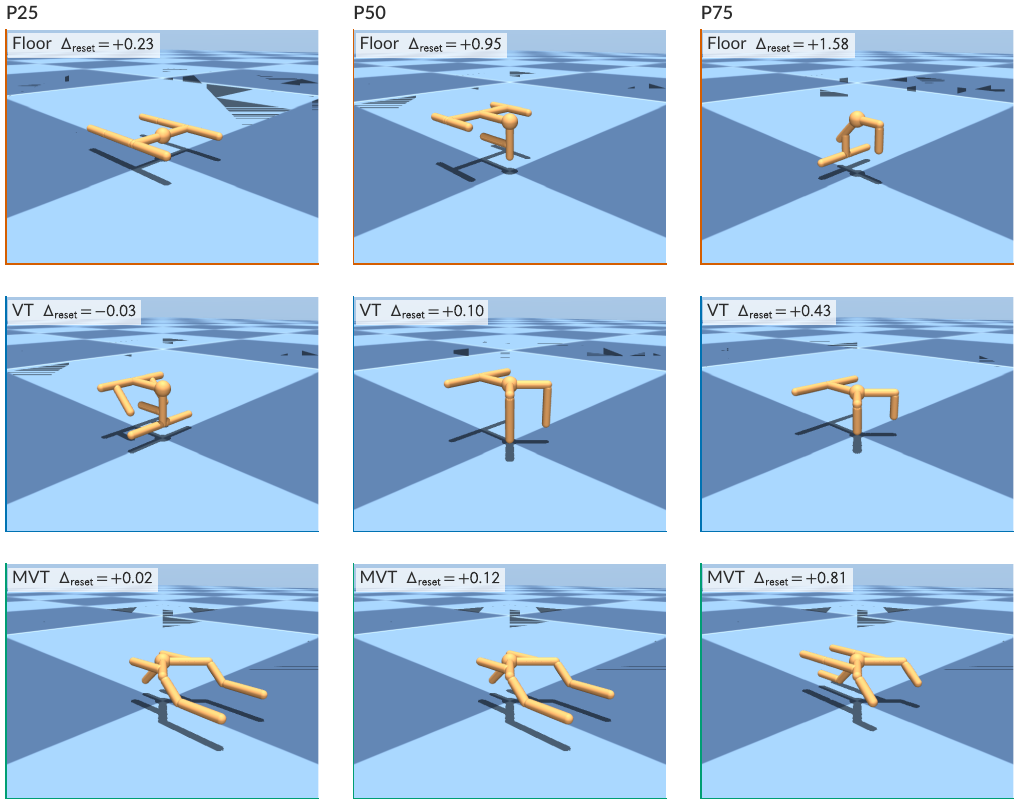}
    \caption{Nine representative UNIMAL morphologies rendered at their reset (spawn) pose. Rows show Floor, VT, and MVT families. Columns show the 25th, 50th, and 75th percentile of $\Delta_{\text{reset}} = \text{PSR}_{\text{Std}} - \text{PSR}_{\text{RF}}$ within each family. For the Floor 25th-percentile body the Standard cutoff lies below the torso radius, and for the MVT 25th- and 50th-percentile bodies it lies 4\,mm above the radius and never ended a recorded evaluation episode, so their $\Delta_{\text{reset}}$ should not be read as an effect of the Standard rule. \textbf{Takeaway:} protocol sensitivity appears across visibly different bodies within each evolutionary family.}
    \label{fig:morph_gallery}
\end{figure}

\section{Rank Agreement Across Protocols and Seeds}
\label{sec:rf_vs_strict_appendix}

\begin{figure}[htbp]
    \centering
    \includegraphics[width=0.95\textwidth]{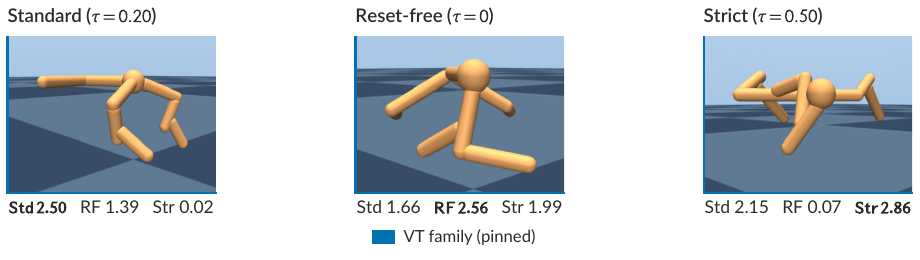}
    \caption{Protocol-specialization exemplars within VT. Family is held fixed, and specialization $=$ PSR$_p$ $-$ max PSR over the other two protocols. Triplet: cross-protocol PSR (bold $=$ training protocol). The Standard and Strict exemplars show sharp cross-protocol collapse (PSR\,$\leq 0.07$), and the Standard exemplar's Strict score of 0.02 comes from Strict runs cut off during the drop from the reset pose (Appendix~\ref{sec:fallrules}). \textbf{Takeaway:} protocol specialization remains visible even within one evolutionary family.}
    \label{fig:top1_vt}
\end{figure}

\paragraph{Leave-one-seed-out reselection for Figure~\ref{fig:top1_vt}.} Re-running the most-protocol-specialized selection within VT preserves the qualitative pattern: the Standard and Strict specialists reappear across folds, and reset-free alternatives remain VT bodies with the same protocol-specialization profile.

\subsection{Main Results on Each Body Set and the Seed-Noise Reference}
\label{sec:ranking_appendix}

Table~\ref{tab:survive} gives the main results on each body set, including a fourth, without the 23 cut-off-seed bodies (Appendix~\ref{sec:fallrules}). The reordering and the larger Reset-free cost of Floor bodies hold on every set, and also when both problems are removed at once, on the 65 of the 72 bodies without a cut-off seed (Reset-free vs.\ Strict $\rho = 0.37$; Floor minus VT on RF/Std $-0.30$, permutation $p = 0.010$, post hoc). The Strict-side family difference does not: the gap between the Floor and VT losses under Strict training is not significant even on all 100 bodies (post hoc permutation $p = 0.25$) and vanishes on the 72, and the low share kept by Standard-only selection comes mostly from the bodies whose cutoffs do not act as intended.

\begin{table}[htbp]
\centering
\caption{Main results on all 100 bodies, on the 72 whose Standard cutoff lies above the torso radius, and without the 11 cut-off or the 23 cut-off-seed bodies (Section~\ref{sec:cutoff_check}). The seed-noise reference is the agreement that seed noise alone would leave between the two three-seed rankings, and $p$-values permute family labels across bodies. Strict loss is one minus Strict retention, and the last row is the in-sample share of the Strict top-5 reward kept by the five best bodies after Standard training. The three right-hand columns are post hoc, as are the seed-noise, $D$, Strict-loss and Standard top-5 rows and the $p$-values in the Floor minus VT and Reset-free vs.\ Standard rows.}
\label{tab:survive}
\footnotesize
\setlength{\tabcolsep}{4pt}
\begin{tabularx}{\linewidth}{@{}Ycccc@{}}
\toprule
Result & All 100 & The 72 & \begin{tabular}[b]{@{}c@{}}Without\\the 11 (89)\end{tabular} & \begin{tabular}[b]{@{}c@{}}Without\\the 23 (77)\end{tabular} \\
\midrule
Reset-free vs.\ Strict $\rho$ & .21 & .26 & .25 & .45 \\
Seed-noise reference for that $\rho$ & .89 & .89 & .85 & .90 \\
Reset-free vs.\ Strict $D$ & .53 & .52 & .42 & .38 \\
Reset-free vs.\ Standard $D$ & .17 & .20 & .16 & .20 \\
RF/Std, Floor/VT/MVT & .55/.84/.72 & .44/.76/.60 & .56/.81/.72 & .47/.80/.71 \\
Floor minus VT on RF/Std ($p$) & $-.30$ ($<$.001) & $-.32$ (.002) & $-.25$ (.008) & $-.33$ (.001) \\
Reset-free vs.\ Standard $\times$ family ($p$) & $<$.001 & .003 & .007 & $<$.001 \\
Three-rule $\times$ family ($p$) & .004 & .01 & .14 & .008 \\
Strict loss, Floor/VT/MVT (\%) & 22/34/9 & 16/17/1 & 19/10/9 & 8/4/6 \\
Standard top 5, \% of Strict top-5 PSR & 49.0 & 75.7 & 72.7 & 95.6 \\
\bottomrule
\end{tabularx}
\end{table}

\paragraph{Seed-noise reference.}
Averaging three seeds makes a ranking more reliable. If $W$ is the mean Spearman correlation between single-seed rankings of one rule, the Spearman--Brown formula $3W/(1 + 2W)$ gives the reliability of a three-seed ranking as 0.93 for Reset-free, 0.90 for Standard and 0.85 for Strict. The formula matches the data: it predicts the agreement between one seed and the mean of the other two at 0.86 for Reset-free and 0.72 for Strict, against 0.86 and 0.73 observed. If seed noise were the only difference between the rules, the three-seed Reset-free and Strict rankings would therefore agree at about $\sqrt{0.85 \times 0.93} = 0.89$ $[0.84, 0.92]$, against the observed 0.21.

\paragraph{Controls for the reordering.}
Scored by the neutral evaluator, which has no cutoff, the Reset-free and Strict rankings of the 100 bodies correlate at $0.31$ $[0.11, 0.49]$. On the 60 bodies whose $0.10\,z_0$ cutoff lies below the torso radius and that no probe rollout pushed below it, training with $\tau = 0.10$ is in effect Reset-free training, run as a separate launch with the same seeds. The two three-seed rankings correlate at $0.92$ $[0.84, 0.96]$, close to the $0.89$ expected with independent seeds, while Reset-free and Strict training of the same bodies correlate at $0.44$ $[0.20, 0.64]$ (difference $0.48$ $[0.27, 0.72]$). Both checks are post hoc.

\paragraph{Excess disagreement.}
For rules $a$ and $b$ with within-rule agreements $W_a$ and $W_b$ and a mean cross-rule correlation $C$ between single-seed rankings, the excess disagreement $D = (W_a + W_b)/2 - C$ is zero in expectation when changing the rule moves a ranking no more than drawing a new seed does. On all 100 bodies, $D$ is 0.17 $[0.08, 0.28]$ for Reset-free versus Standard, 0.29 $[0.17, 0.43]$ for Standard versus Strict and 0.53 $[0.36, 0.70]$ for Reset-free versus Strict, with percentile intervals that resample bodies. A null that shuffles each body's Reset-free and Strict runs between two pseudo-rules centers at 0.64, with 95\% of draws between 0.55 and 0.73, again far above the observed 0.21.

\paragraph{Top-$k$ overlap.}
Seeds alone turn over the top of a ranking, so we use top-$k$ overlap as a description rather than a test. Two single-seed runs of the same protocol share 3 to 8 of their top-10 bodies, and two simulated three-seed reruns of one rule share about 7 of their top-10 bodies, so the observed 5 of 10 has a probability of 0.03 to 0.15 under seed noise alone. The top-20 lists of Reset-free and Strict training share 10 bodies, against 15.5 to 16.1 in simulated seed reruns and 13.7 under the exchangeability null ($p = 0.001$). For top-$k$ overlap intervals, the bootstrap holds one protocol's observed top-$k$ fixed.

\FloatBarrier

\section{Locomotion-Quality Audit}
\label{sec:locomotion_quality_appendix}

Low per-step reward under Reset-free training could reflect a body that \emph{learned to walk but was penalized by the protocol}, or a body that \emph{never learned to walk at all}. We compare the distance traveled after each training protocol under the common and the neutral evaluator.

We classify a (morphology, training protocol) cell as \emph{walks} if the mean evaluation distance is at least $5\,\text{m}$ and as \emph{stationary} if below $1\,\text{m}$. We report cross-protocol transitions at both $\tau_{\text{eval}} = 0.20$ (the main-text evaluation) and $\tau_{\text{eval}} = 0$ (no early termination, so distance is integrated over the full evaluation horizon), in Tables~\ref{tab:locomotion_quality} and~\ref{tab:locomotion_quality_neutral}.

\begin{table}[H]
\centering
\caption{Per-family cross-protocol locomotion transitions at $\tau_{\text{eval}} = 0.20$. A body that walks under Standard but is stationary under the target protocol either learned locomotion and was disrupted by the protocol or, for the Floor-reset-free cell, learned a policy whose evaluation episodes the fall cutoff often ends early.}
\label{tab:locomotion_quality}
\small
\begin{tabular}{lccccc}
\toprule
Family & $n$ & Std$\to$RF halt & Std$\to$Str halt & Std halt$\to$RF walks & Std halt$\to$Str walks \\
\midrule
Floor & 41 & 12 & 1  & 0 & 0 \\
VT    & 37 & 4  & 10 & 0 & 0 \\
MVT   & 22 & 5  & 0  & 0 & 0 \\
\bottomrule
\end{tabular}
\end{table}

\begin{table}[H]
\centering
\caption{Per-family cross-protocol locomotion transitions at $\tau_{\text{eval}} = 0$ (no early termination; distance integrated over the full evaluation horizon). The VT-Strict and MVT-Strict counts hold unchanged, but all ten VT-Strict stops are bodies with cut-off Strict runs, so they are not a family effect (Appendix~\ref{sec:fallrules}).}
\label{tab:locomotion_quality_neutral}
\small
\begin{tabular}{lccccc}
\toprule
Family & $n$ & Std$\to$RF halt & Std$\to$Str halt & Std halt$\to$RF walks & Std halt$\to$Str walks \\
\midrule
Floor & 41 & 0  & 1  & 0 & 0 \\
VT    & 37 & 0  & 10 & 0 & 0 \\
MVT   & 22 & 0  & 0  & 0 & 0 \\
\bottomrule
\end{tabular}
\end{table}

Twelve of 41 Floor bodies walk under Standard but become stationary under reset-free at $\tau_{\text{eval}} = 0.20$, compared to only one that becomes stationary under strict; four VT bodies become stationary under reset-free, ten under strict; MVT shows five vs.\ zero. Under $\tau_{\text{eval}} = 0$ with no early termination, none of the 12 Floor bodies stays under 1\,m after reset-free training, and 5 of the 12 walk at least 5\,m, and the VT-Strict and MVT-Strict counts are unchanged (10 and 0). All ten VT-Strict stops are bodies with cut-off Strict runs, so they are not a family effect (Appendix~\ref{sec:fallrules}). This audit supports attributing the family ordering to the training reset rule: under neutral evaluation, RF/Std ratios remain ordered Floor $0.643 <$ MVT $0.847 <$ VT $0.953$ (Figure~\ref{fig:robustness_eval}), and fixed-horizon fitness ratios similarly keep Floor lowest (Appendix~\ref{sec:per_seed_metric}).

\FloatBarrier
\section{Control-Effort Confound Audit}
\label{sec:pareto_appendix}

We derive $\widehat{\sum_t \|a_t\|^2}_{\text{per episode}} = (\text{distance}/dt - \text{fitness})/10^{-3}$ with $dt = 0.02$\,s (this recovers each run's mean per-episode sum from its recorded mean distance and return, given the reward in Appendix~\ref{sec:appendix_env}). Per-step $\|a_t\|^2$ is family-protocol-flat at $8.8$--$9.8$, so the family difference in Section~\ref{sec:results} is not explained by one family pushing actuators harder. Among runs whose controllers travel at least $5\,$m per episode, the squared action norm per meter (not torque, work, or physical energy) is $178$--$266$ across all nine family$\times$protocol cells; differences do not mirror the PSR family-by-protocol interaction.

\section{Extended Discussion}
\label{sec:extended_discussion_appendix}

\paragraph{Implications for autonomous reset-free RL.}
Reset-free RL~\citep{sharma2022autonomous, yang2025morefree, patil2024risc, gupta2021mtrf} removes episodic resets; our $\tau{=}0$ condition removes one component of the reset schedule, the height-triggered early termination. The body-dependent sensitivity we observe suggests a concrete test for broader reset-free settings: if morphology changes rankings under matched finite-horizon training, algorithms developed for settings with fewer resets~\citep{sharma2022autonomous, yang2025morefree, patil2024risc} should be evaluated across body families rather than on a single fixed morphology. Floor-family bodies are candidate stress tests for recovery from difficult states. This motivates co-designing bodies and algorithms under the reset schedule of interest.

\paragraph{Why sweep the termination threshold.}
Co-design and locomotion implementations configure the height-termination threshold in different ways (Appendix~\ref{sec:protocol_glossary_appendix}), so the body rankings they report can depend on a choice that the audited sources do not sweep as a benchmark variable. Removing the height predicate during training, our $\tau{=}0$ condition, is also a finite-horizon analogue of what an autonomous-RL pipeline would face if it inherited a co-design-selected body and had to learn without hand-tuned fall detectors. The $\tau{=}0$ condition is a co-design-relevant corner of the protocol space; the paper's claims are scoped to the simulator (cf.\ Section~\ref{sec:limitations}).

\FloatBarrier
\section{Time Limit and Other Protocol Variables}
\label{sec:boundary_scale_appendix}

Section~\ref{sec:boundary_free} reports that, by its planned criterion, the no-boundary variant changes the ranking ($\rho = 0.85$, 95\% CI $[0.67, 0.93]$, whose upper end lies below the planned bound of 0.95) and family magnitudes relative to corpus $\tau{=}0$. Its runs used a later code revision than the corpus. A panel of 12 bodies retrained with that revision under all three rules (108 runs) ranks the bodies as the corpus does at $\rho = 0.97$, 0.88 and 0.73 for Reset-free, Standard and Strict, and in a post hoc comparison the revision lowers Reset-free Floor scores by a similar amount as the no-time-limit panel (a median of 0.42 PSR on that panel's 4 Floor bodies and 0.48 on 7 Floor bodies of a longer-budget panel, both with the ordinary 1,000-step limit), so the panel does not isolate the time limit.

\begin{figure}[!htbp]
    \centering
    \includegraphics[width=\textwidth]{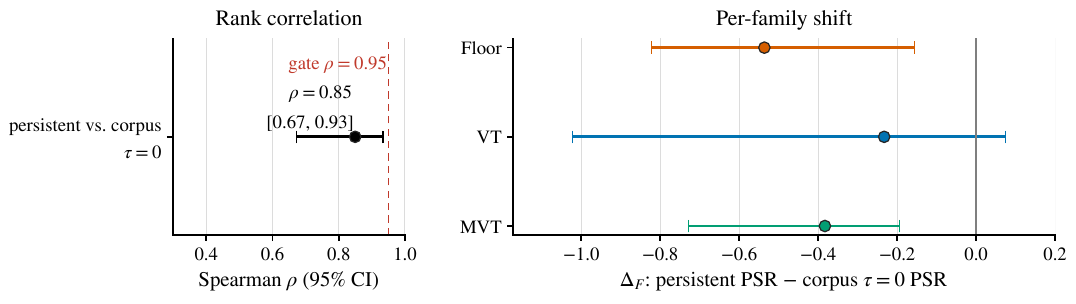}
    \caption{Persistent no-boundary PPO vs.\ corpus $\tau{=}0$ on a 30-body $\times$ 3-seed panel, trained with a later code revision than the corpus. Left: per-body rank correlation with corpus $\tau{=}0$ is $\rho = 0.85$, and the upper end of its interval lies below the planned gate of 0.95. Right: family-level shifts (persistent minus corpus), median $\Delta_F$ with 95\% bootstrap CIs. \textbf{Takeaway:} the no-time-limit runs, trained with later code, rank the bodies differently from the corpus and score Floor and MVT bodies lower.}
    \label{fig:boundary_free_forest}
\end{figure}

A planned factorial test on the same 30 bodies, which also trained Standard and Strict without a time limit (180 runs), finds that the reset rule and the time limit interact ($F(2, 505) = 3.72$, $p = 0.025$), which meets the plan's criterion. Additional boundary-duration panels give the same qualitative reading. Spearman correlations against the corpus $B{=}1000$ setting are $\rho_{3001}=0.916$ (95\% CI $[0.784,0.971]$), $\rho_{10001}=0.852$ ($[0.680,0.945]$), and $\rho_{5\text{M}}=0.850$ ($[0.672,0.934]$). The point estimates decrease as the boundary lengthens, while confidence intervals overlap. A per-body regression of $\Delta(B)=\mathrm{PSR}(1000)-\mathrm{PSR}(B)$ on $\log B$ has slope $0.099$ (95\% CI $[0.053,0.150]$), consistent with larger magnitude shifts under longer boundaries. These runs used the same later code. The no-boundary panel is therefore a boundary-setting check for the corpus reset-free condition, not an isolated test of the time limit.

\subsection{Other Design Spaces and Protocol Variables}
\label{sec:otherspaces_appendix}

We also trained panels in other design spaces and on other protocol variables, each with a report-only decision rule. In three pools of 30 parametric Humanoid bodies trained with Brax PPO under four fall rules (none, the native Gymnasium height range, and cutoffs at 0.2 and 0.5 of the root height) and scored by one evaluator without fall termination, rankings under a fall rule agree with rankings without one at 0.58 to 0.73 in two pools and 0.11 to 0.19 in the third. Seed halves of one rule also agree poorly (0.26 to 0.69), and although $D$ is above zero in the third pool (0.28 $[0.01, 0.55]$ for the native rule), no $D$ stays above zero after adjustment for multiple comparisons, so this panel cannot resolve an effect of the fall rule beyond seed variation. Changing how time limits are handled (Humanoid and Evolution Gym), how resets are initialized, or adding observation noise (Ant) left rankings within the agreement between seeds ($\rho = 0.65$ to 0.99). The terminal Humanoid arm used later code than its bootstrapped arm, and much of the Ant agreement comes from a wide spread of skill. In these checks, reordering beyond seed variation appears only for fall termination on UNIMAL bodies, so whether another benchmark is sensitive has to be checked rather than assumed.

\FloatBarrier
\Needspace{0.80\textheight}%
\section{Companion Chronophotograph: Bodies that Walk Under All Three Protocols}
\label{sec:chronophoto_appendix}

\begin{figure}[H]
    \centering
    \includegraphics[width=\textwidth]{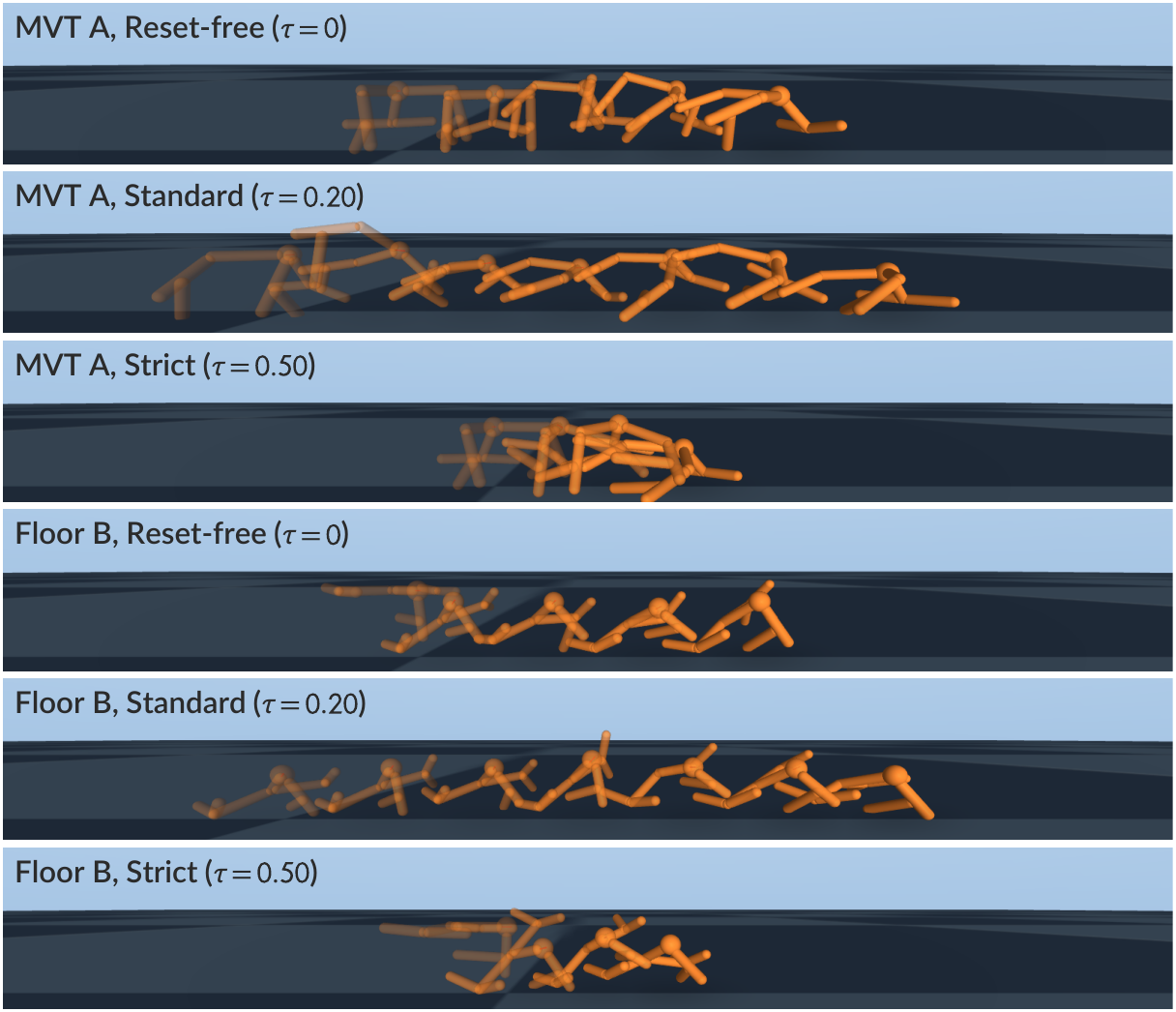}
    \caption{3D-rendered chronophotograph of \emph{MVT A} and \emph{Floor B} from illustrative retrains across three training protocols. Each panel overlays poses from one rollout seen by a fixed camera, so their spread shows the distance covered. \textbf{Takeaway:} protocol dependence is body-specific, as \emph{MVT A} and \emph{Floor B} walk under all three protocols.}
    \label{fig:gait_chronophoto_full}
\end{figure}

Figure~\ref{fig:gait_chronophoto_full} shows \emph{MVT A} and \emph{Floor B} under the three protocols: both walk across all three, as in all 9 recorded corpus runs of each.

\end{document}